\documentclass[letterpaper]{article} 
\usepackage[submission]{aaai25}  
\usepackage{times}  
\usepackage{helvet}  
\usepackage{courier}  
\usepackage[hyphens]{url}  
\usepackage{graphicx} 
\DeclareGraphicsExtensions{.pdf,.png,.jpg,.jpeg}  
\usepackage{natbib}  
\usepackage{caption} 
\usepackage{amsmath}
\usepackage{amssymb}
\usepackage{multirow}
\usepackage[table]{xcolor}
\usepackage{algorithm}
\usepackage{algorithmic}
\usepackage{newfloat}
\usepackage{listings}
\DeclareCaptionStyle{ruled}{labelfont=normalfont,labelsep=colon,strut=off} 
\floatstyle{ruled}
\newfloat{listing}{tb}{lst}{}
\floatname{listing}{Listing}
\title{ShotVL: Human-centric Highlight Frame Retrieval via Language Queries}
\author {
    Wangyu Xue\equalcontrib\textsuperscript{\rm 1},
    Chen Qian\equalcontrib\textsuperscript{\rm 1},
    Jiayi Wu\textsuperscript{\rm 2},
    Yang Zhou\textsuperscript{\rm 2},
    Wentao Liu\textsuperscript{\rm 2},
    Ju Ren\protect\thanks{Corresponding Author, renju@tsinghua.edu.cn}\textsuperscript{\rm 1},
    Siming Fan\protect\thanks{Corresponding Author, gzfansiming@gmail.com}\textsuperscript{\rm 2},
    Yaoxue Zhang\textsuperscript{\rm 1}
}
\affiliations {
    \textsuperscript{\rm 1}Department of Computer Science and Technology, Tsinghua University\\
    \textsuperscript{\rm 2}SenseTime Research\\
}

\usepackage{bibentry}

\begin{document}
\maketitle
\begin{abstract}

Existing works on human-centric video understanding typically focus on analyzing specific moment or entire videos. However, many applications require higher precision at the frame level. In this work, we propose a novel task, BestShot, which aims to locate highlight frames within human-centric videos via language queries. This task demands not only a deep semantic comprehension of human actions but also precise temporal localization. To support this task, we introduce the BestShot Benchmark. 
The benchmark is meticulously constructed by combining 
human-annotated highlight frames, detailed textual descriptions and duration labeling. 
These descriptions encompass three critical elements: (1) Visual content; (2) Fine-grained action; and (3) Human Pose Description. Together, these elements provide the necessary precision to identify the exact highlight frames in videos.

To tackle this problem, we have collected two distinct datasets: (i) ShotGPT4o Dataset, which is algorithmically generated by GPT-4o and 
(ii) Image-SMPLText Dataset, a 
dataset with large-scale and accurate per-frame pose description leveraging PoseScript and existing pose estimation datasets. Based on these datasets, we present a strong baseline model, ShotVL, fine-tuned from InternVL, specifically for BestShot. 
We highlight the impressive zero-shot capabilities of our model and offer comparative analyses with existing SOTA models. ShotVL demonstrates a significant 52\% improvement over InternVL on the BestShot Benchmark and a notable 57\% improvement on the THUMOS14 Benchmark, all while maintaining the SOTA performance in general image classification and retrieval.
\end{abstract}
\section{Introduction}
\begin{figure*}[!t]
\centering
\includegraphics[width=0.68\textwidth]{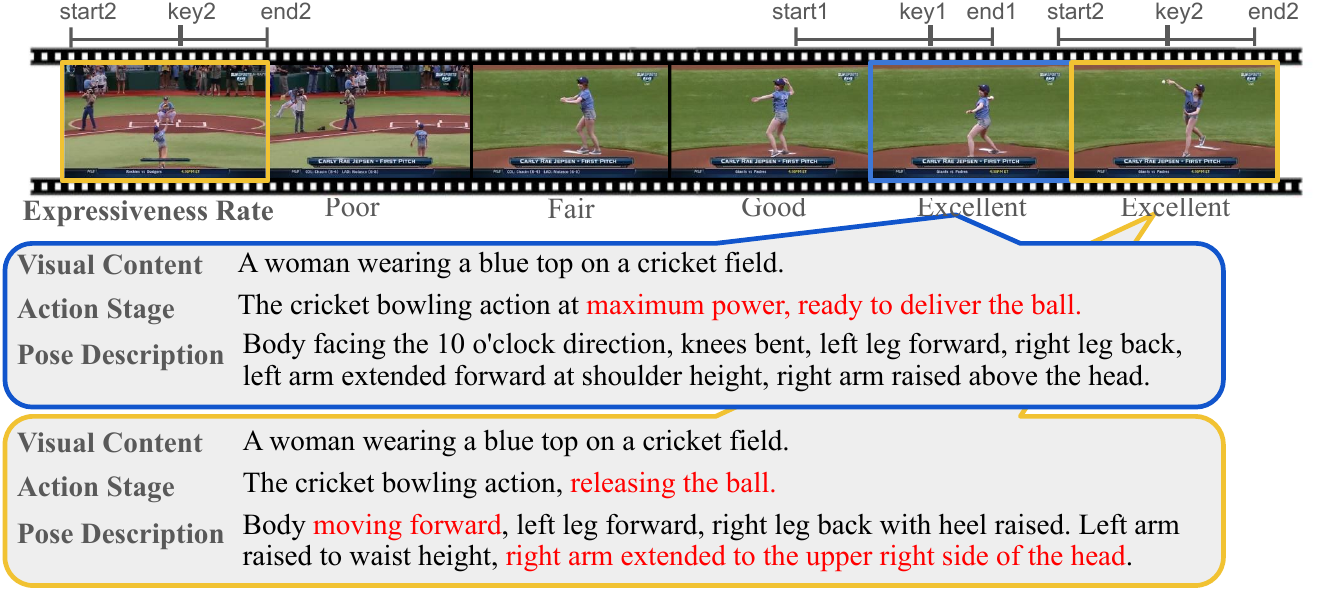} 
\caption{Example of BestShot Benchmark. The task is to locate the exact frame through language queries related to content, action stage and pose description. Each query may correspond to multiple intervals.}
\label{fig1}
\end{figure*}
\begin{table}[!ht]
\centering
\setlength{\tabcolsep}{1mm} 
\begin{tabular}{ccccccc}
\hline
\textbf{Dataset}  & \textbf{Domain}                                         & \textbf{\begin{tabular}[c]{@{}c@{}}\#Qu\\ ery\end{tabular}} & \textbf{\begin{tabular}[c]{@{}c@{}}Dura-\\ tion\end{tabular}} & \textbf{\begin{tabular}[c]{@{}c@{}}\#Ac-\\ tion\end{tabular}} & \textbf{\begin{tabular}[c]{@{}c@{}}\#Vi-\\ deo\end{tabular}} & \textbf{\begin{tabular}[c]{@{}c@{}}Avg \\ word\end{tabular}} \\ \hline
AVA-2.2           & Movie                                                   & 80                                                          & 1s                                                            & -                                                             & 430                                                          & 2                                                            \\
THUMOS14              & Sports                                                  & 20                                                          & 4.3s                                                          & 20                                                            & 413                                                          & 3                                                            \\
FAction           & Daily                                                   & 106                                                         & 7.1s                                                          & 106                                                           & 16.7K                                                        & 2                                                            \\
MSports           & Sports                                                  & 66                                                          & 1.0s                                                          & 4                                                             & 0.8K                                                         & 2                                                            \\
HiRest            & Daily                                                   & 8K                                                          & 7.6s                                                          & -                                                             & 3.4K                                                         & 4                                                            \\
DiDeMo            & Flickr                                                  & 41K                                                         & 8.0s                                                          & -                                                             & 10.6K                                                        & 7                                                            \\
ANet              & Activity                                                & 72K                                                         & 7.2s                                                          & -                                                             & 15K                                                          & 14                                                           \\
Charades          & Activity                                                & 16K                                                         & 13.4s                                                         & -                                                             & 6.7K                                                         & 6                                                            \\
QVHL              & \begin{tabular}[c]{@{}c@{}}Vlog/\\ News\end{tabular}    & 10K                                                         & 11.3s                                                         & -                                                             & 10.2K                                                        & 11                                                           \\
\textbf{BestShot} & \begin{tabular}[c]{@{}c@{}}TH14,\\ FAction\end{tabular} & 6K                                                          & \textbf{\begin{tabular}[c]{@{}c@{}}12\\ frame\end{tabular}}   & 59                                                            & 628                                                          & 78                    \\ \hline                                      
\end{tabular}
\caption{Benchmark and Dataset Comparison: TH14(THUMOS 14), FAction (FineAction), MSports (MultiSports). BestShot is only for zero-shot test.}
\label{tabdata}
\end{table}

Within the domain of video understanding, locating highlight frames in human-centric videos using language queries is a crucial yet under-explored task. Many downstream tasks rely heavily on high-precision frame understanding to function effectively. For video step captioning, it is vital for generating accurate video captions that align with specific actions and timestamps, such as in sports (``a player scores a goal between 00:01:05 and 00:01:08") or in instructional videos (``the chef flips the pancake at exactly 00:02:03"). In moment retrieval, precise frame localization ensures that the retrieved moments match the query exactly, particularly for short-duration or complex actions.



Despite the importance of precise frame-level localization, existing methods face significant limitations in meeting these demands, primarily stemming from insufficient frame-level granularity and lack of diverse frame data. For instance, many current approaches provide only broad temporal segments rather than precise frame-level annotations, limiting their effectiveness in tasks that require detailed understanding. Another problem lies in the restricted length of their training queries, usually ten words or fewer, which hinders the retrieval of frames with complex or nuanced descriptions. Furthermore, these methods frequently struggle to generalize across diverse scenes and actions, diminishing their reliability in real-life applications where content is often more varied and unpredictable.

In this paper, we introduce a challenging benchmark, the BestShot Benchmark. This benchmark aims to address the limitations of existing tasks and benchmarks by providing a more granular evaluation metric that better reflects the demands of high-precision frame understanding. With the BestShot Benchmark, it becomes more convenient and reliable to evaluate models'ability to accurately localize highlight frames in human-centric videos using detailed queries.


The BestShot Benchmark highlights several key obstacles that must be overcome: (1) Lack of relevant data: First, existing large-scale image and video captioning datasets emphasize coarse-grained summaries rather than detailed descriptions. Moreover, fine-grained annotations, such as those found in ShareGPT4V~\cite{chen2023sharegpt4v}, are limited both in scale and precision, particularly when it comes to capturing detailed action stages and pose descriptions. Lastly, most moment retrieval datasets are manually annotated and relatively small, restricting their scalability and the generalization ability of models trained on them. (2) Limitations of Existing Methods: The performance of current baseline methods on the proposed benchmark remains subpar. Large vision-based models, such as InternVL~\cite{chen2023internvl} and InternVideo~\cite{wang2022internvideo}, struggle to localize frames in videos without fine-tuning on domain-specific data. Additionally, while video LLMs like LITA~\cite{huang2024lita} and VTG-LLM~\cite{guo2024vtg} possess some localization and temporal understanding capabilities, their generalization is hindered by the insufficiency of training data.

\begin{figure}[h]
\centering
\includegraphics[width=0.58\linewidth]
{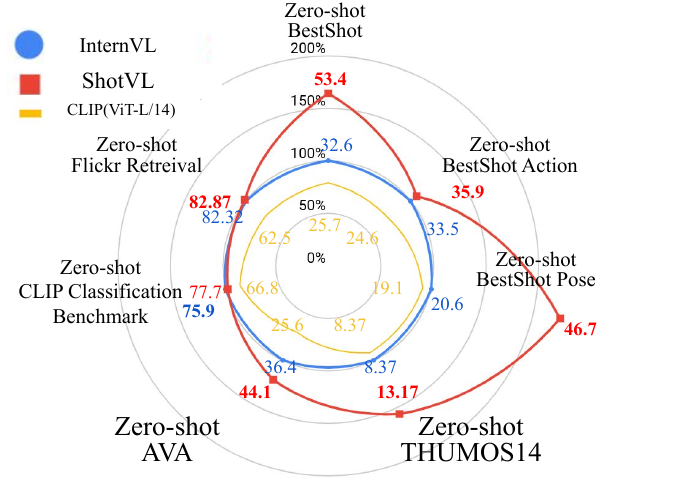} 
\caption{Zero-shot evaluation on BestShot, Temporal Action Localization (THUMOS14), Action Classification (AVA), and CLIP Classification and Retrieval Benchmark.}
\label{fig:radar}
\end{figure}

Therefore, following the benchmark, we also propose two large-scale training dataset relevant to BestShot and a robust baseline ShotVL for tackling these challenges. Our main contributions can be summarized as follows:
\begin{itemize}
\item  We introduce the BestShot Benchmark for precise highlight frame retrieval in human-centric videos using complex and fine-grained language queries.
\item We propose the large-scale ShotGPT4o Dataset and ImageSMPLText Dataset to address the issues of limited data diversity and insufficient fine-grained descriptions.
\item We provide a comprehensive evaluation of existing methods and propose a robust baseline ShotVL. ShotVL demonstrates a significant 52\% improvement over InternVL on the BestShot Benchmark and a notable 57\% improvement on the THUMOS14 Temporal Action Localization Benchmark~\cite{THUMOS14}, all while maintaining the SOTA performance in general image classification and retrieval as shown in Fig.\ref{fig:radar}\footnote{We use both AVA Actions and AVA-Kinetics and these two datasets are short for AVA in this paper.}.
\end{itemize}

\section{Related Work}
\subsection{Datasets and Tasks}
\begin{itemize} 
\item{\textbf{Shot by Image aesthetics.}}
A straightforward approach to BestShot leverages Image Quality Assessment (IQA) and Image Aesthetics Assessment (IAA) techniques, such as those in the AVA dataset~\cite{murray2012ava}, to select the highest-scoring frame. In practice, frame scoring is often combined with query-based retrieval to optimize frame selection.

\item{\textbf{Shot by Query}}
Query-based moment retrieval includes Temporal Action Localization (TAL) and Moment Retrieval (MR) task. TAL datasets includes THUMOS14, FineAction, MultiSports, and ActivityNet and so on. These are limited by fixed queries and models trained on those lack zero-shot capabilities. MR datasets like Charades-STA, QVHighlight, HiRest~\cite{zala2023hierarchical}, QueryD~\cite{oncescu2021queryd}, and DiDeMo~\cite{anne2017localizing} involve annotated queries but focus more on long segments, leaning towards general video understanding rather than highlight frame retrieval in human-centric videos.

\item{\textbf{Shot by Pose}}
Using pose descriptions for highlight retrieval is underexplored. PoseScript~\cite{delmas2022posescript} is the first large-scale pose description dataset, though it only annotates AMASS SMPL~\cite{AMASS:ICCV:2019}, not real images. Both Motion-X~\cite{lin2024motionxlargescale3dexpressive} and our work utilize a PoseScript-like annotation method to describe poses in real-world videos. We additionally validated that this approach is highly effective for locating pose-query frames in real videos. Such general pose description meets some highlight frame retrieval needs and is more user-friendly compared to dividing each action into dozens of specialized sub-actions.

\item{\textbf{Shot by Video Question \& Answer (Q\&A)}}
Activitynet-RTL~\cite{huang2024lita}, Time-IT~\cite{ren2024timechat}, and VTG-IT~\cite{guo2024vtg} convert moment retrieval annotations into video Q\&As formats using GPT, without adding new localization data. They then fine-tune LLMs on their proposed dataset, enabling LLMs to acquire moment retrieval capabilities in specific domains.
\end{itemize}

\subsection{Vision Language Models}

\begin{itemize} 
\item{\textbf{Image-based.}}
Although CLIP and its derivative methods 
CoCa~\cite{yu2022cocacontrastivecaptionersimagetext}, InternVL) can directly achieve frame retrieval in videos via language queries, they struggle with distinguishing subtle differences between adjacent frames and are insensitive to long queries describing detailed information such as fine-grained action stages and poses. Visual LLMs using the above base models like  
InternVL-Chat~\cite{chen2023internvl,chen2024fargpt4vclosinggap} face similar issues. ChatHuman~\cite{lin2024chathuman}, which ensembles multiple models by LLM, and PoseGPT~\cite{posegpt}, which regresses SMPL using LLM, are more related to our work since both focus on human-centric images.
\item{\textbf{Video-based.}}
Although InternVideo performs well on supervised Temporal Action Localization tasks, demonstrating strong temporal understanding, it falls short in understanding fine-grained, frame-level details and zero-shot retrieval within video, compared to CLIP and InternVL, which are trained only on image-text pairs. Current Video LLMs capable of retrieval within video, such as TimeChat, VTG-LLM, and LITA, often use image vision base models, like CLIP, as the vision encoder rather than video base models. On the other hand, video LLMs built on video vision base models (e.g. InternVideo) like VideoChat~\cite{li2024videochatchatcentricvideounderstanding}, and VideoGPT+~\cite{maaz2024videogptintegratingimagevideo} have yet to be validated for their localization capabilities in retrieval within video.
\end{itemize}

\section{Benchmark and Training Data}
In this section, we discuss the annotation methods for the 
 BestShot Benchmark (Fig. \ref{fig1}) and training data for the proposed baseline model ShotVL.

\begin{figure*}[t]
\centering
\includegraphics[width=0.68\textwidth]{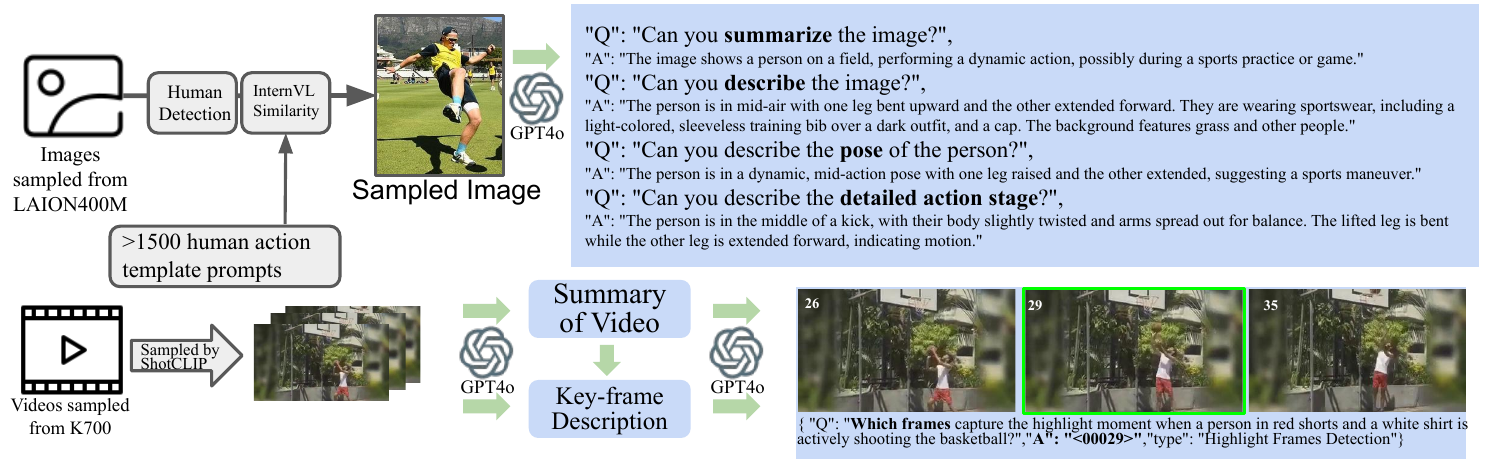} 
\caption{\textbf{Annotation Pipeline of ShotGPT4o}}
\label{fig:preS3}
\end{figure*}

\subsection{Human-annotated Zero-shot Benchmark}
To construct the benchmark, We randomly selected 10\% of videos covering 60 action categories from THUMOS14 and FineAction for re-annotation. All ground truths in our benchmark are human-annotated, following a three-step pipeline:

\begin{enumerate}
    \item \textbf{Selecting Potential Highlight Frames}: Identify potential highlight frames from the sequences. Two annotator groups score each frame on human expressiveness (considering action importance, pose extent, and facial expression, similar to Image Aesthetic Assessment but more objective), with scores from 1 to 5. Frames averaging above 4.5 are considered potential highlights.
    \item \textbf{Write detail queries for Highlight Frames}:  
Annotators were asked to write a detailed description of the highlight frame that uniquely identifies it within the video. Descriptions are divided into three parts: (1) \textbf{Visual Contents}: Distinguish the main character from background characters, (2) \textbf{Fine-Grained Action Stages}: Describe key moments in actions in detail, e.g. "peak of jump" instead of "person is jumping", (3) \textbf{Significant Human Poses}: annotate salient poses during actions using general descriptions. This is essential for professional actions like gymnastics and complex actions like dances.
    \item \textbf{Label Start-End Segments}: Define start-end frames in the video based on the text descriptions to ensure accuracy of ground truth query. 
\end{enumerate}

\textbf{Metrics.}
The Top@1 retrieval accuracy metric was used, where a prediction is correct if the predicted frame falls within any of the ground-truth intervals in the video. We divided 6,000 queries into three categories: Content, Action, and Pose, with 2,000 queries each. The ``Full" metric combines all three queries into one, ``Action" focuses only on action queries, and ``Pose" focuses only on pose queries. For Pose, the ground-truth interval allows a tolerance of four frames around the key frame, while Content and Action use manually annotated intervals with an average of 12 frames.

\subsection{Hybrid Training Data and Annotation Pipeline}
We ensured three criteria to avoid benchmark leak when constructing the training data: (1) The annotations we proposed are entirely generated by GPT or automatically synthesized, without human annotation\footnote{Except for 2,000 human-written pose descriptions, crucial for aligning SMPLText, GPTPoseText, and Human-written PoseText. Such training data cause benchmark leak, thus we presents a further ablation study in Table \ref{tab:stage1abla} related to this dataset.}. (2) We did not sample frames or actions from the THUMOS14 or FineAction datasets. (3) Frames were randomly selected from the LAION-400M~\cite{schuhmann2021laion400mopendatasetclipfiltered} datasets, without focusing on the 60 action categories in the benchmark.


\textbf{ShotGPT4o Dataset:}
We used GPT4o to annotate image captions with tolly 600K frames. It consist (1) 200K images and 800K captions, (2) another 150K images and 700K image Q\&As, (3)250K frames descriptions and video Q\&As from 10K videos and corresponding 100K video Q\&As including multiple task like frame retrieval, moment retrieval, dense captioning, video summary and 
visual reasoning. The images are sampled from LAION-400M and videos are sampled from K700, covering at least 1,500 action categories as shown in Fig.~\ref{fig:preS3}.  Note that to avoid benchmark leak of K700 to better evaluate the zero-shot performance, the video Q\&As annotations are not used in experiment Tab.~\ref{tab:stage1result} and Tab.~\ref{tab:stage1abla}, but used in the video chat application in Fig.~\ref{fig:videochat}.
Unlike prior work~\cite{liu2023llava,chen2023sharegpt4v,chen2015microsoftcococaptionsdata} focused solely on summary or fine-grained descriptions, we split Q\&As into four dimensions: summary, description, action and detail action stage, and pose description. The annotation pipeline is shown in Fig. \ref{fig:preS3}(a). For video Q\&A, we also propose a novel annotation strategy modified from the ShareGPT4Video~\cite{chen2024sharegpt4video}, to further improve the accuracy of frame description and and frame or moment retrieval Q\&As. As this annotation strategy and its videos Q\&As is not crucial to train ShotVL. It is just an initial attempt at frame retrieval using temporal cues, and the effectiveness is shown in the Fig.~\ref{fig:preS3} and Fig.~\ref{fig:videochat}. More details will be described in the supplementary materials.

\begin{figure}[!t]
\centering
\includegraphics[width=0.87\linewidth]{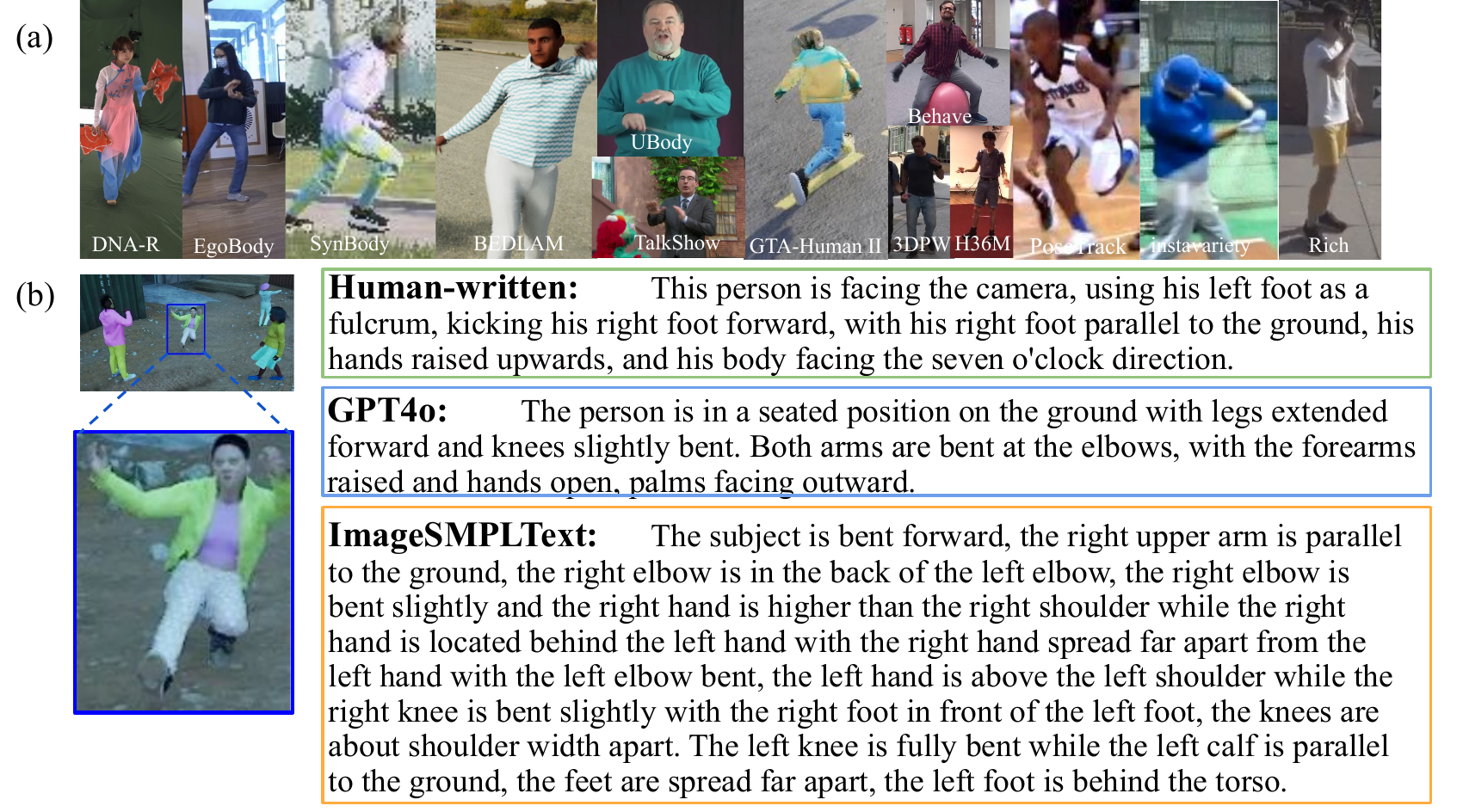} 
\caption{(a) Example datasets with ground-truth SMPL annotation we used. (b) Difference of pose descriptions among Human-written, GPT-4o and Image-SMPLText.}
\label{fig:smplexample}
\end{figure}

\textbf{Image-SMPLText Dataset:}
BestShot Benchmark improvements were limited by the high noise level in GPT-4o's 375K pose descriptions. To address this, we introduce the Image-SMPLText Dataset (Fig.~\ref{fig:smplexample}), adapting PoseScript to real-world videos to re-annotate over 13 public video datasets~\cite{yi2023generating,vonMarcard2018,h36m_pami,andriluka2018posetrackbenchmarkhumanpose,lin2023osx,Huang:CVPR:2022,zhang2022egobody,cai2021playing,humanMotionKZFM19,yang2023synbodysyntheticdatasetlayered,2023dnarendering}.
This generated 18.6M pose descriptions based on SMPL joint positions and orientations. We enhanced PoseScript with body orientation to better suit real-world videos. 
Additionally, 2,000 samples were manually described to align SMPLText, pose description in ShotGPT4o, and human-written descriptions. While Motion-X also generates SMPLText, our focus is on real videos with accurate SMPL ground truth, unlike Motion-X's use of crawled data. As shown in Tab.~\ref{tab:smpldata} and Fig.~\ref{fig:smplexample}, we only use the video dataset with accurate SMPL ground-truths which have been proven effective in SMPLer-X~\cite{cai2024smplerxscalingexpressivehuman}, as crawled video data introduces noise that can negatively impact training. Fig.~\ref{fig:smplexample} illustrates the differences between human-written descriptions, GPT-4o, and Image-SMPLText.
Fig.~\ref{fig:smplexample}(b) shows an example of difference between human-written, GPT-4o and Image-SMPLText.

\begin{table}[!t]\small
\setlength{\tabcolsep}{1mm} 
\centering
\begin{tabular}{llllll}
\multicolumn{3}{c}{\textbf{Motion-X}}                 & \multicolumn{3}{c}{\textbf{Ours}}                     \\ \hline
\textbf{Dataset} & \textbf{\#Clip} & \textbf{\#Frame} & \textbf{Dataset} & \textbf{\#Clip} & \textbf{\#Frame} \\ \hline
Aist++           & 1470            & 340K             & BEDLAM           & 10000           & 951K             \\
Animation$^{*2}$      & 329             & 38K              & SynBody          & 6K              & 633K             \\
Dance$^{*2}$          & 163             & 36K              & InstaVariety     & 14K             & 2.1M             \\
Egobody          & 980             & 438K             & GTA-H II         & 10K             & 1.8M             \\
Fitness$^{*2}$        & 16.7K           & 358K             & EgoBody          & 125             & 935k             \\
Game$^{*2}$           & 10.2K           & 1.1M             & RICH             & 496             & 243K             \\
GRAB$^{*1}$           & 1.3K            & 406k             & UBody            & 836             & 683k             \\
HAA500         & 5.2K            & 311K             & PoseTrack        & 1.3K            & 28K              \\
Humanml$^{*1}$        & 26K             & 3.5M             & BEHAVE           & 299             & 44K              \\
Humman           & 744             & 104K             & H3.6M            & 840             & 312K             \\
Idea400          & 12.5K           & 2.5M             & 3DPW             & 61              & 22.7K            \\
kungfu$^{*2}$         & 1040            & 257K             & DNA-R            & 24K             & 5M               \\
perform$^{*2}$        & 475             & 102K             & TalkShow         & 984             & 3.3M           \\ \hline 
\end{tabular}
\caption{Components of Public Datasets. *1 indicates the dataset have not images, *2 indicates the dataset is from the Internet and may have no accurate SMPL. As a Comparison, Motion-X dataset does not entirely consist of images, and a significant portion is sourced from videos, leading to issues with inaccurate keypoint predictions, which in turn result in imprecise pose descriptions. In contrast, we only use publicly available video datasets that include accurate SMPL ground truth. By comparison, PoseScript describes only the AMASS dataset, which lacks real images.}
\label{tab:smpldata}
\end{table}

\subsection{Analysis \& Limitation of Benchmark \& Data}
\textbf{Zero-shot BestShot benchmark.}
Tab~\ref{tabdata} shows the detail of BestShot Benchmark. It is collected from 112K THUMOS14 frames and 2M FineAction frames, including 400 THUMOS14 clips and 500 FineAction videos. Fig.~\ref{fig1} shows two query samples. To validate the benchmark's accuracy, we had another group of annotators locate key frames in the video based on the query. The average human retrieval rate was 86\%, significantly surpassing current SOTA methods.

\textbf{ShotGPT4o.}
Manual validation of 500 Q\&As showed that only 70\% of pose descriptions are accurate; GPT-4o often confuses left and right. This lower accuracy explains the poor results when training solely on GPT-4o's pose descriptions. Additionally, while 90\% of action descriptions are correct, only 22\% are fine-grained. For instance, 78\% of throwing actions are described generally as ``extending the body during a throw," with only 22\% detailing stages like ``about to release the ball," ``just released the ball," or ``at maximum power." This explains why the BestShot Benchmark's action dimension improves by only 10\% with GPT-4o data.
\textbf{Image-SMPLText.}
There still remains some problems.
 The first is the lack of temporal infomation, which is partly solved in PoseFix~\cite{delmas2024posefixcorrecting3dhuman} and MotionScript~\cite{yazdian2023motionscriptnaturallanguagedescriptions}. They re-wrote pose-code in PoseScript to generate caption of two frames and a few sequences. We also generated more than 18.6M sequences description in this way, but not yet used in training. 
 The second is the insufficient pose-code related to real-world videos. Although we have addressed the lack of global orientation code, while pose information related to translation and velocities is not finished yet. 
 The third problem is that the unseen body part is also described in detail since the pose caption is automatically generated from ground-truth SMPL. We try to use the confidence from keypoint prediction, which is mentioned in Motion-X, to solve this problem, that is to assume that keypoints with lower confidence are occluded, which is described as ``... is occluded" or just not to be described. However, experiments show that the confidence cannot accurately reflect occlusion in most cases. Considering that model has the ability to guess the pose of occluded body part, we ultimately decided not to introduce the concept of occlusion in the final implementation.

\begin{figure*}[t]
\centering
\includegraphics[width=0.68\textwidth]{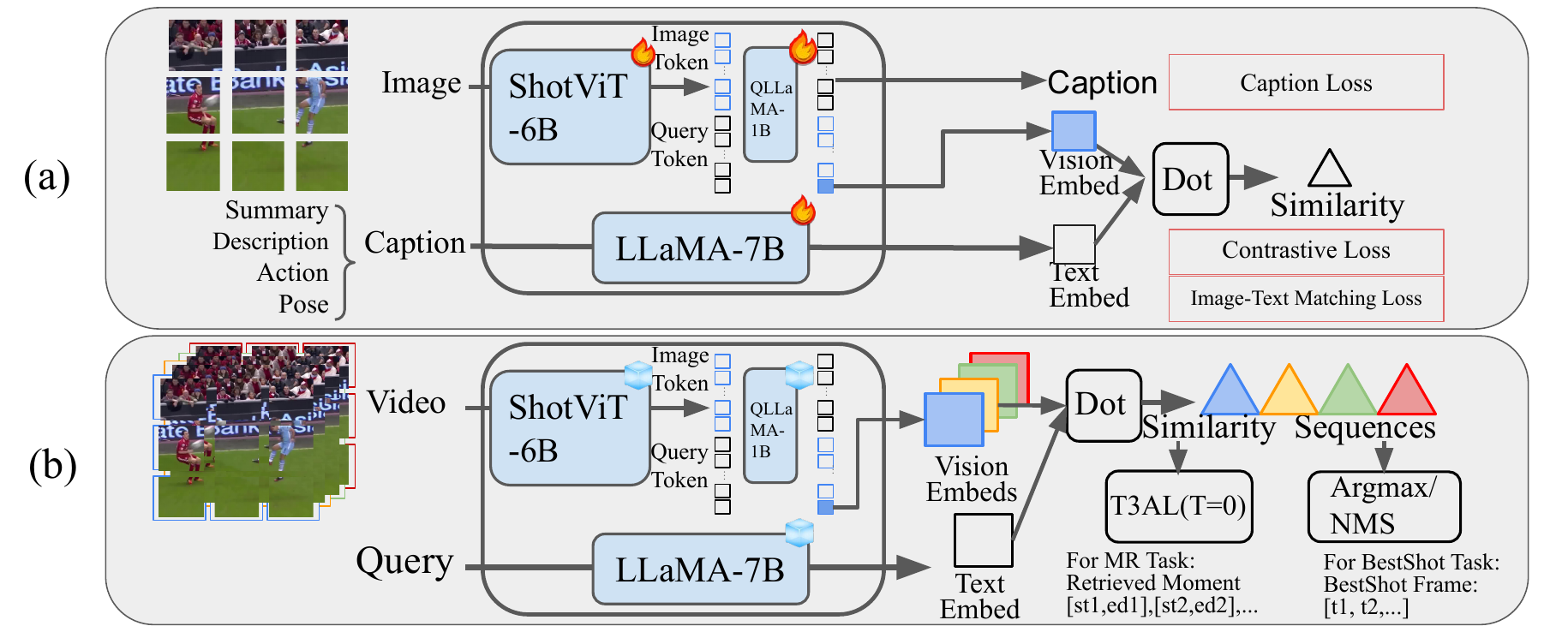} 
\caption{Training  and inference pipeline of ShotVL. (a) Training. (b) Inference of BestShot and Moment Retrieval. }
\label{fig:train}
\end{figure*}

\section{Method}
In this section, we present the proposed baseline model, ShotVL (Fig.~\ref{fig:train}). Given a video and text query, 
ShotVL is able to locate the most relevant frames or moments. To address the challenges in the BestShot task, we trained ShotVL following the fine-tune stage of InternVL-Base model. The key designs used in ShotVL are listed as follows:




    \textbf{Base Model Selection}. 
    Although InternVideo excels at Supervised Temporal Action Localization tasks like THUMOS14 and FineAction, the BestShot task demands fine-grained image understanding and very short sequence localization. Thus, InternVideo, pre-trained only on video-text pairs, underperforms compared to the smaller CLIP model. And InternVL significantly outperforms CLIP in ZeroShot BestShot tasks, leading us to select it as our baseline model.

    \textbf{Furthest Point Sampling}
    During training, we applied furthest point sampling (FPS) on 18.6M Image-SMPLText samples, creating 10 different 1\% and 2\% subsets from varying start frames to avoid impact of excessive similarity between adjacent frames. First, we randomly chose a frame as the start frame and calculated mean per-joint position error (MPJPE) after orientation alignment with other frames. The frame with the largest MPJPE would be selected as a new start, and we would repeat this process until we sampled the required number of frames from the whole dataset.
    
    \textbf{Customized Data and Data Ratios}
    Due to the significant differences between Image-SMPLText and the original training data of InternVL,  
   training only on SMPLText weakens the pre-trained model's performance on general image tasks. To counter this, we added the COCO dataset~\cite{chen2015microsoftcococaptionsdata}, a high-quality human-written image caption dataset, and adjusted data ratios to maintain general abilities. We also replaced some Image-SMPL-Text descriptions with GPT-4o-generated or human-written captions to bridge the style gap between dataset descriptions and general queries.



\textbf{Inference Pipeline} During inference, frame-by-frame similarities with queries are computed first. For the BestShot task, simply argmax to get the best frame, or apply non-maximum suppression (NMS) to get predicted frames. For moment retrieval or temporal action localization task, we follow T3AL (T=0)~\cite{Liberatori_2024_CVPR}, which first uses mean of video vision feature to match a pseudo-label from the list of possible actions, then segment the sequences according to similarities.
\begin{table*}[t]\small
\setlength{\tabcolsep}{1mm}
\centering
\begin{tabular}{lc|ccc|c|c|ccc}

\multicolumn{1}{c}{}       &       & \multicolumn{3}{c|}{\begin{tabular}[c]{@{}c@{}}Zero-shot\\ BestShot\end{tabular}}                                                                                       & \begin{tabular}[c]{@{}c@{}}Zero-shot\\ TAL\end{tabular}  & \begin{tabular}[c]{@{}c@{}}Zero-shot\\ AC\end{tabular}             & \multicolumn{3}{c}{\begin{tabular}[c]{@{}c@{}}Zero-shot\\ CLIP Benchmark\end{tabular}}                                                                                               \\ \hline
\multicolumn{1}{c}{method} & size  & \begin{tabular}[c]{@{}c@{}}Full\\ (top1)\end{tabular} & \begin{tabular}[c]{@{}c@{}}Action\\ (top1)\end{tabular} & \begin{tabular}[c]{@{}c@{}}Pose\\ (top1)\end{tabular} & \begin{tabular}[c]{@{}c@{}}THUMOS14\\ (avg mAP)\end{tabular} & \begin{tabular}[c]{@{}c@{}}AVA\\ mean (acc1/5)\end{tabular} & \begin{tabular}[c]{@{}c@{}}flickr,I2T,\\ (top1)\end{tabular} & \begin{tabular}[c]{@{}c@{}}flickr,T2I\\ (top1)\end{tabular} & \begin{tabular}[c]{@{}c@{}}avg(C)\\ (top1)\end{tabular} \\ \hline
(a1)CLIP ViT-B/32          & 188M  & 24.6                                                  & 19.1                                                    & 7.6                                                   & -                                                        & 22.52                                                              & 73.75                                                        & 55.38                                                       & 53.78                                                   \\
(a2)CLIP ViT-L/14          & 406M  & 25.7                                                  & 24.6                                                    & 19.1                                                  & 7.65                                                    & 25.57                                                              & 81.45                                                        & 62.50                                                        & 66.75                                                   \\
(a3)LongCLIP               & 406M  & 31.3                                                     & 30.1                                                       & 21.2                                                     & 7.41                                                        & 35.22                                                                  & 86.35                                                            & 74.28                                                           & 68.95                                                       \\
(a4)InternVL          & 14B   & 32.6                                                  & 33.5                                                    & 20.6                                                  & 8.37                                                     & 36.42                                                              & 92.40                                                        & 82.32                                                       & \textbf{77.70}                                          \\ \hline
(b)InternVideo1           & 1.3B  & 21.6                                                  & 23.8                                                    & 17.7                                                  & 4.00                                                        & *                                                                  & ×                                                            & ×                                                           & ×                                                       \\ \hline
(c1)VTG-LLM                & 8.4B  & 6.2                                            & 7.6                                              & 5.1                                            & 0.66                                                     & ×                                                                  & ×                                                            & ×                                                           & ×                                                       \\
(c2)LITA                   & 13.2B & 12.6                                                 & 12.2                                                   & 14.1                                                 & 0.17                                                     & ×                                                                  & ×                                                            & ×                                                           & ×                                                       \\ \hline
\textbf{ShotVL}                & 14B   & \textbf{53.4}                                         & \textbf{35.9}                                           & \textbf{46.7}                                         & \textbf{13.17}                                          & \textbf{44.12}                                                     & \textbf{93.05}                                               & \textbf{82.87}                                              & 75.90                                                  
\end{tabular}
 \caption{Quantitative comparison of SOTA models on zero-shot BestShot Benchmark, THUMOS14 Temporal Action Localization (TAL), AVA Action Classification (AC), CLIP Retrieval Benchmark (only Flickr~\cite{young2014image} evaluated; COCO excluded due to benchmark leak during training; I2T and T2I denote image-to-text and text-to-image), and CLIP Classification Benchmark (average of 25 tasks). TAL testing uses the T3AL(T=0) method for the base model. *InternVideo's zero-shot AVA-Kinetic~\cite{li2020avakineticslocalizedhumanactions}  result is not evaluated due to prior training on Kinetics. (a) SOTA base models with single-frame input, (b) SOTA base models with multi-frame input, (c) SOTA video LLMs with multi-frame input and segment output, we use the middle frame of the predicted interval for frame prediction, as it yields the highest score compared to other frames. We evaluated top1 accuracy in BestShot, mean of mAP under iou={0.3,0,4,0.5,0.6,0.7} in THUMOS14, mean of top1 and top5 accuracy in AVA, and top1 accuracy in CLIP Benchmark.}
\label{tab:stage1result}
\end{table*}

\section{Experiment}
Quantitative comparison of SOTA models on zero-shot BestShot Benchmark, THUMOS14 Temporal Action Localization, AVA Action Classification, CLIP Retrieval Benchmark (only Flickr evaluated; COCO excluded due to benchmark leak during our training; I2T and T2I denote image-to-text and text-to-image), and CLIP Classification Benchmark (average of 20 tasks). TAL testing uses the T3AL(T=0) method~\cite{Liberatori_2024_CVPR} for the base model. 

\begin{table*}[!t]\small
\setlength{\tabcolsep}{1mm}
\centering
\begin{tabular}{l|ccccc|ccc|cc|ccc}
\multicolumn{1}{c|}{}       & \multicolumn{5}{c|}{Finetuned Data}                                                                                                                                                    & \multicolumn{3}{c|}{\begin{tabular}[c]{@{}c@{}}Zero-shot\\ BestShot\end{tabular}} & \begin{tabular}[c]{@{}c@{}}Zero-shot\\ TAL\end{tabular} & \begin{tabular}[c]{@{}c@{}}Zero-shot\\ AC\end{tabular} & \multicolumn{3}{c}{\begin{tabular}[c]{@{}c@{}}Zero-shot\\ CLIP Benchmark\end{tabular}} \\ \hline
\multicolumn{1}{c|}{method} & \begin{tabular}[c]{@{}c@{}}Human\\ Pose\end{tabular} & \begin{tabular}[c]{@{}c@{}}GPT\\ Pose\end{tabular} & \begin{tabular}[c]{@{}c@{}}SMPL\\ Pose\end{tabular} & \begin{tabular}[c]{@{}c@{}}Shot\\ GPT4o\end{tabular}     & General & Full                      & Action                    & Pose                      & THUMOS14                                                    & AVA-K                                                  & flickr,I2T                 & flickr,T2I                  & avg(C)                      \\ \hline
(a1)InternVL                & N                                                    & N                                                  & N                                                   & N          & N       & 32.6                      & 33.5                      & 20.6                      & 8.37                                                    & 36.42                                                  & 92.40                       & 82.32                       & 77.70                       \\
(a2)InternVL-ft           & Y                                                    & N                                                  & N                                                   & N          & N       & 40.5                      & 35.2                      & 28.9                      & 9.59                                                    & 42.7                                                   & 92.45                      & 82.15                       & \textbf{77.78}              \\ \hline
(b1)ShotVL-PoseS           & N                                                    & N                                                  & \textbf{Y}                                          & N          & N       & 40.8                      & 28.1                      & 33.2                      & 8.25                                                    & 36.14                                                  & 74.40                       & 56.62                       & 74.33                       \\
(b2)ShotVL-PoseSH           & \textbf{Y}                                           & N                                                  & \textbf{Y}                                          & N          & N       & 48.6                      & 27.9                      & 42.7                      & 7.22                                                    & 46.06                                                  & 77.20                       & 59.36                       & 74.09                       \\
(b3)ShotVL-Pose           & \textbf{Y}                                           & \textbf{Y}                                         & \textbf{Y}                                          & N          & N       & 49.4                      & 24.1                      & 43.8                      & 5.34                                                    & \textbf{47.83}                                         & 77.20                       & 59.36                       & 74.09                       \\ \hline
(c1)ShotVL-G           & \textbf{Y}                                           & \textbf{Y}                                         & \textbf{Y}                                          & N          & LAION   & 48.9                      & 29.5                      & 39.5                      & 7.92                                                    & 42.99                                                  & 88.65                      & 75.99                       & 76.85                       \\
                            & \textbf{Y}                                           & \textbf{Y}                                         & \textbf{Y}                                          & N          & COCO    & 51.5                      & 33.1                      & 42.4                      & 9.90                                                    & 46.33                                                  & \textbf{94.20}              & \textbf{84.05}              & 76.51                       \\ \hline
(d)ShotVL-Full              & \textbf{Y}                                           & \textbf{Y}                                         & \textbf{Y}                                          & \textbf{Y} & COCO    & \textbf{53.4}             & \textbf{35.9}             & \textbf{46.7}             & \textbf{12.49}                                          & 44.41                                                  & 93.05                      & 82.87                       & 75.90                      
\end{tabular}
\caption{Ablation Study on Finetuned Data and Model Performance for InternVL and ShotVL(We rename ShotVL for the model trained on our proposed ShotGPT4o and Image-SMPLText datasets). Human/GPT Pose (Human-written and GPT-generated pose descriptions), SMPL Pose (our Image-SMPLText dataset), General (image captioning datasets such as LAION400M and COCO captions), Zero-shot TAL (Zero-shot Temporal Action Localization), and Zero-shot AC (Zero-shot Action Classification). Metrics are the same as Tab.~\ref{tab:stage1abla}.}
\label{tab:stage1abla}
\end{table*}

\textbf{Implementation of ShotVL}
ShotVL followed the InternVL's fine-tuning pipeline with our specially-designed datasets selection and ratios. We used InternVL 14B as our base model. As mentioned before, the datasets we used can be divided into 4 categories and these datasets follows the ratios as Image-SMPLText:ShotGPT4o:General=1:5:5, which makes a stable balance between the BestShot task and general retrieval and classification tasks. The ShotVL model has been trained for 2,000 iterations, with batch size 1,536 and learning rate 1e-5, on 24 A100 for 20 hours.

\begin{figure}[!t]
\centering
\includegraphics[width=0.63\linewidth]{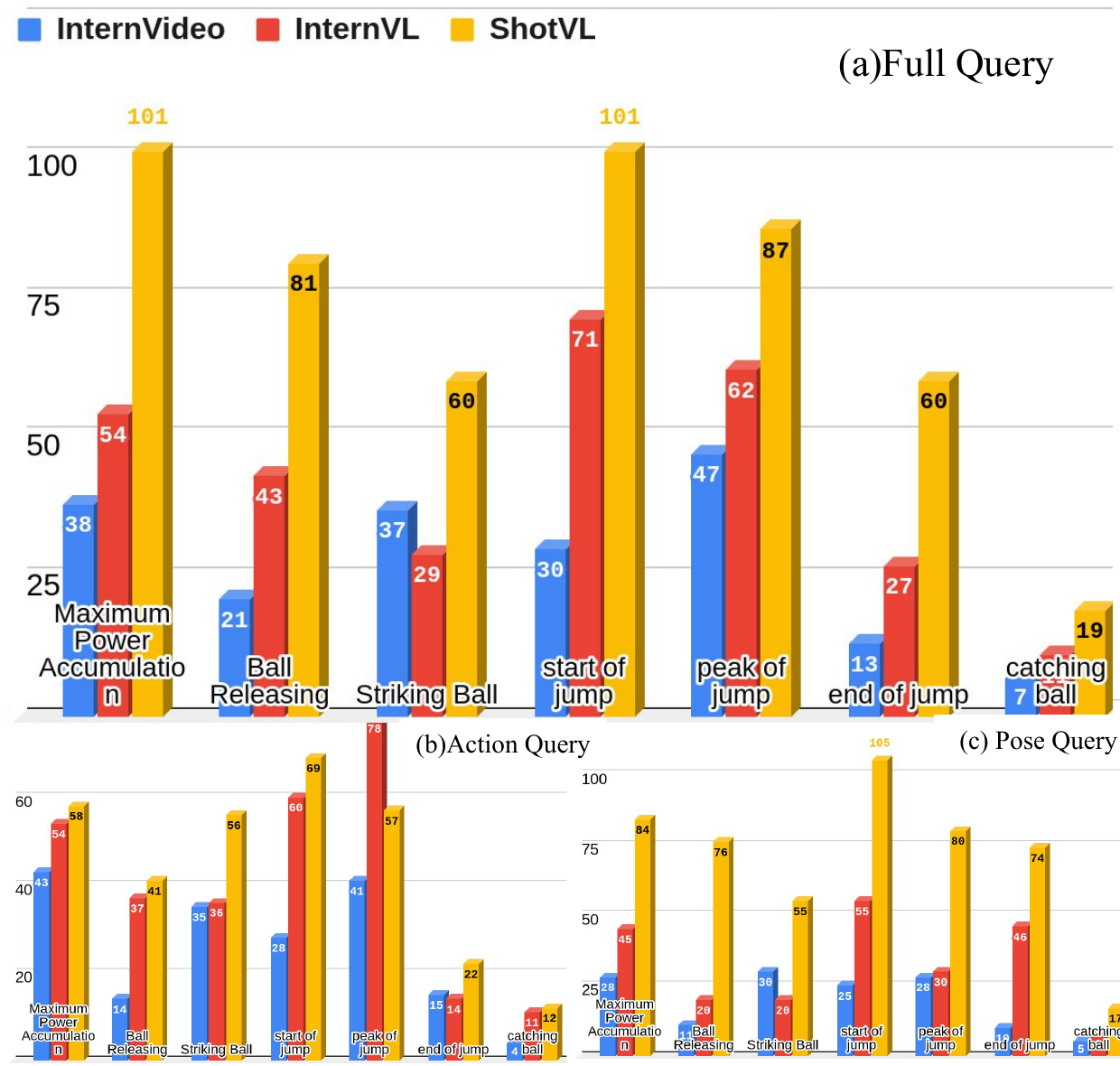} 
\caption{Hit count comparisons on the most common fine-grained action categories on BestShot Benchmark.}
\label{fig:hitcount}
\end{figure}

\textbf{Evaluation of ShotVL}
We tested several SOTA models and ShotVL for zero-shot abilities in frame localization (BestShot), temporal localization (THUMOS14 TAL), action classification (AVA-Kinetics), and CLIP general retrieval and classification benchmarks. As shown in the Tab.~\ref{tab:stage1result}, benchmark metrics improve significantly with model size. CLIP faces challenges in fine-grained long-text retrieval on BestShot due to limited training text length. While LongCLIP~\cite{zhang2024longclip} and interpolation of positional encoding offer partial solutions, progress is hindered by a lack of long-text training data. InternVideo, despite large-scale video-text training and stronger performance on temporal understanding, loses some frame-level spatial details, making a drop in performance. InternVL excels in fine-grained understanding without CLIP’s long-text constraints, with simple fine-tuning yielding significant BestShot improvements. Our ShotVL surpasses InternVL across all metrics, matching InternVL in general CLIP retrieval and classification tasks. We further divided the queries in THUMOS14 subset of BestShot into several fine-grained action categories and listed the accuracy of each model in the most common categories, as shown in Fig.~\ref{fig:hitcount}.


\textbf{Ablation Study}
 Ablation experiments are divided into four groups (Tab.~\ref{tab:stage1abla}). From (a1) to (a2), InternVL was fine-tuned using 2,000 human-annotated pose descriptions. From (b1) to (b3), we conducted ablation on the pose description training data, verifying the effectiveness of Image-SMPLText and the necessity of aligning the three types of pose descriptions. However, the introduction of Image-SMPLText led to a significant decline in general retrieval capabilities. Therefore, starting from experiment (c), we incorporated general image-text data for training, aiming to maintain both pose retrieval and general capabilities. Finally, experiment (d) validated the effectiveness of ShotGPT4o.

\begin{figure}[t]
\centering
\includegraphics[width=0.7\linewidth]{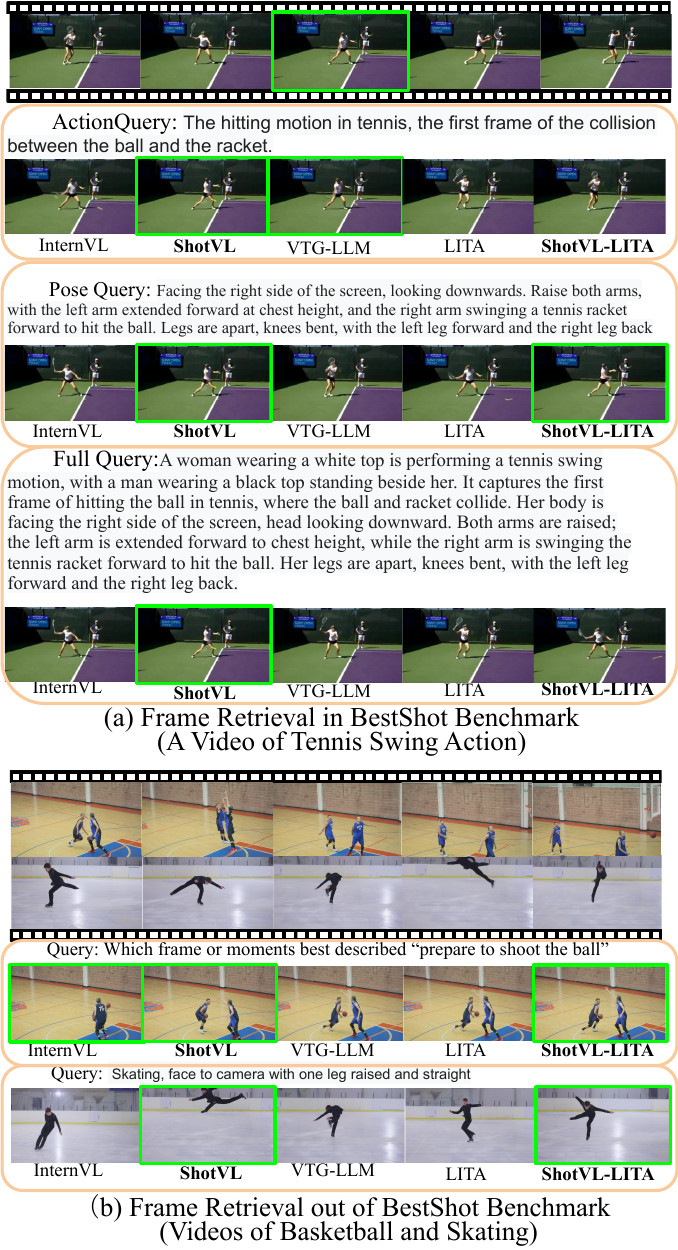} 
\caption{Qualitative comparisons of BestShot Application. }
\label{fig:qualitative_comparison}
\end{figure}

\textbf{Generalization of BestShot.}
 Fig.~\ref{fig:qualitative_comparison}(a) shows the qualitative comparisons of various models under BestShot Benchmark. For queries and scenes that are not shown in the evaluated benchmarks, 
Fig.~\ref{fig:qualitative_comparison}(b) demonstrated ShotVL's strong generalization ability.
\begin{figure}[h]
\centering
\includegraphics[width=0.7\linewidth]{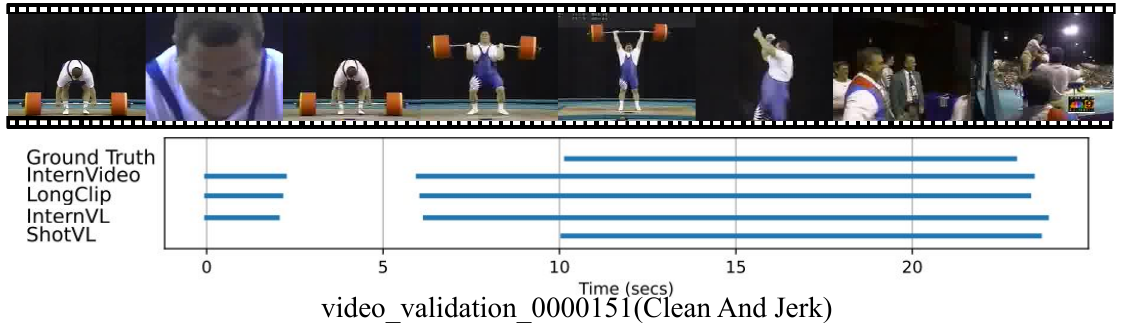} 
\caption{Qualitative comparisons of Zero-shot TAL. }
\label{fig:tal}
\end{figure}

\begin{figure}[!ht]
\centering
\includegraphics[width=0.7\linewidth]{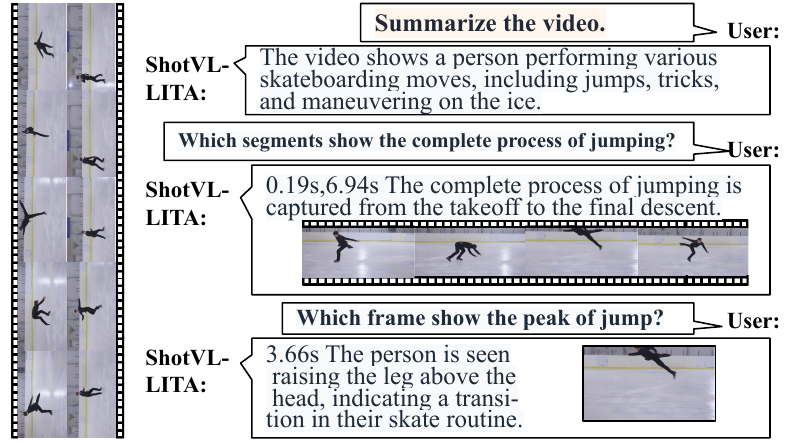} 
\caption{An example of video chat application.}
\label{fig:videochat}
\end{figure}


\textbf{Zero-shot Temporal Action Localization.}
CLIP, InternVL, and ShotVL can be directly applied to zero-shot temporal action localization via T3AL($T=0$) method. The comparisons are shown in Tab.~\ref{fig:qualitative_comparison} and Fig.~\ref{fig:radar} and Fig.~\ref{fig:tal}.

\textbf{Video Chat.}
ShotVL can be further integrated with a Video LLM, such as LITA or VTG-LLM, to have capabilities for video question answering, moment retrieval, and frame retrieval at the same time as shown in Fig.~\ref{fig:videochat}.
We adapt LITA as the baseline method to train our Video Chat model. The modifications include: (1) replace encoder to ShotVL, (2) pre-train and fine-tune on image Q\&As following LLaVA to connect ShotVL with Vicuna. (3) add frame token into LLM, (4) fine-tune on multiple video instruction data including ShotGPT4o.



\section{Conclusion}

We collected the BestShot Benchmark, a highlight-frame-retrieval-in-video benchmark that includes 6,000 queries with manually identified highlight frames, detailed descriptions, and matched temporal segments. We further propose two 
datasets tailored for zero-shot BestShot Task: 
ShotGPT4o and Image-SMPLText.
which have been proven effective on zero-shot BestShot, AVA action classification, and THUMOS14 action localization benchmarks.


Methodologically, we evaluated several SOTA models on the BestShot task and proposed ShotVL as a robust solution. ShotVL shows superior zero-shot performance on BestShot task, yet it remains limited in retrieving actions that require strong temporal correlations. As highlighted in the previous section, a potential future direction is to extend ShotVL by integrating it with Video LLMs and collect a large-scale frame or moment retrieval video dataset to train such model.


\section{Appendices}

In this supplementary material, we provide detail discussion of $(1)$ BestShot Benchmark; $(2)$ ShotGPT4o Dataset; $(3)$ Image-SMPLText Dataset; $(4)$ ShotVL Model; $(5)$ ShotVL-LITA Model.
\section{BestShot Benchmark}

\textbf{Comparisons with existing benchmark}
MultiSports~\cite{li2021multisports} and QVHighLight~\cite{lei2021detecting} are the two datasets most similar to BestShot among existing datasets, and they have comparable data scales. The primary distinction between BestShot and MultiSports is that BestShot supports evaluation for arbitrary queries across 60 action classes, whereas MultiSports only supports a limited number of queries across 4 action classes. The main differences between BestShot and QVHighLight are: 
\begin{itemize}
    \item \textbf{Human-centric}: QVHighLight focuses on Vlogs and News Videos, while BestShot focuses on Human-centric Videos related to sports and human activities.
    \item \textbf{Shorter Moments}: BestShot features frame-level annotations, with queries averaging just 12 frames in duration, whereas QVHighLight provides second-level annotations, with an average query duration of 11.3 seconds.
    \item \textbf{Complex Fine-grained Query}: The average length of a full query in BestShot is 78 words, compared to 11 words in QVHighLight. This setup is designed to further narrow the annotation interval, allowing it to be more distinct from other frames in the video, especially adjacent frames.
    \item \textbf{Saliency Score of Expressiveness}: QVHighLight provides second-level saliency scores, while BestShot offers frame-level expressiveness scores.
\end{itemize}

\begin{figure}[!t]
\centering
\includegraphics[width=0.95\linewidth]{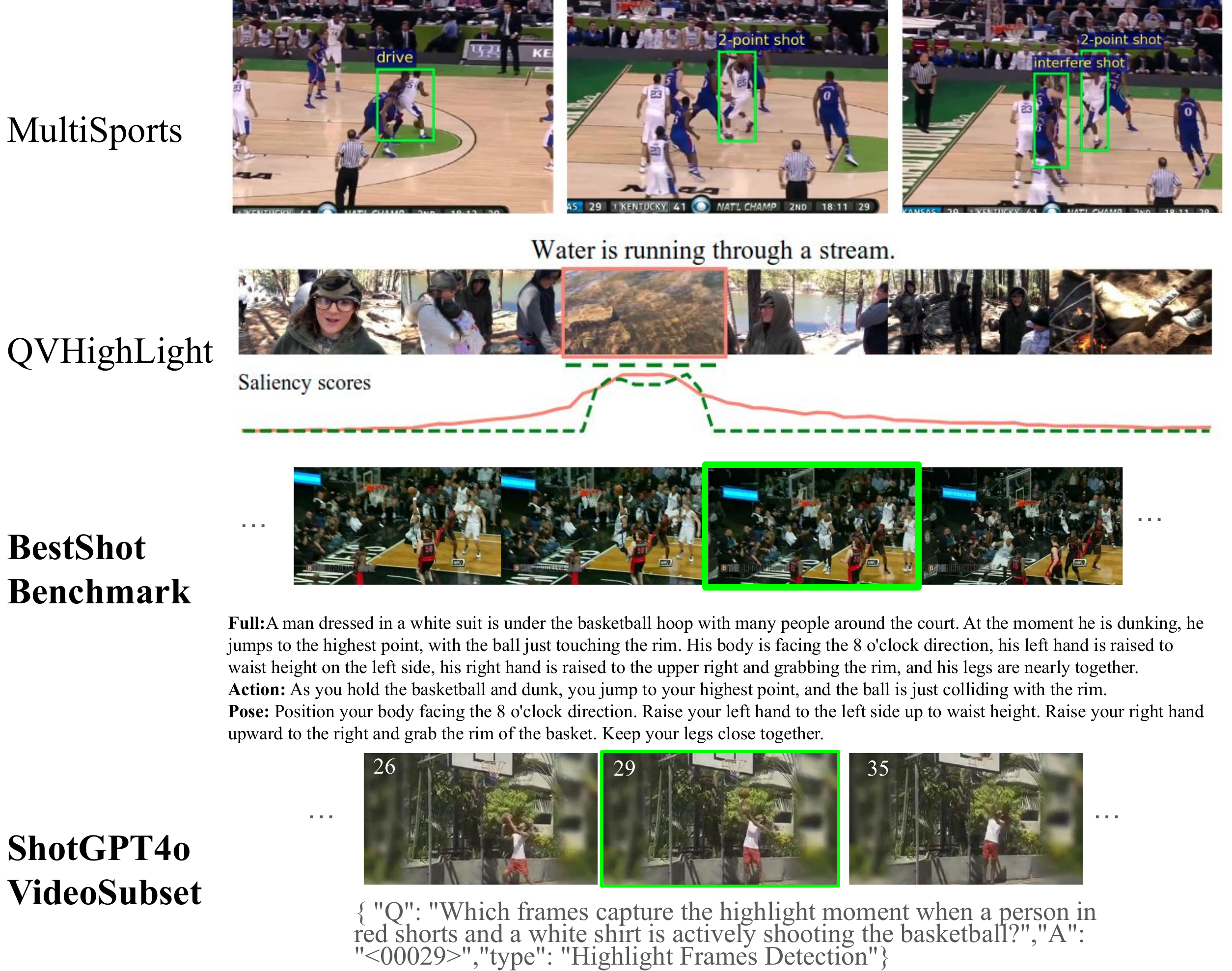} 
\caption{\textbf{Benchmark and dataset comparisons.} Images from the first two line is picked from their paper.}
\label{fig:BestShotExample1}
\end{figure}

\begin{figure*}[t]
\centering
\includegraphics[width=0.95\textwidth]{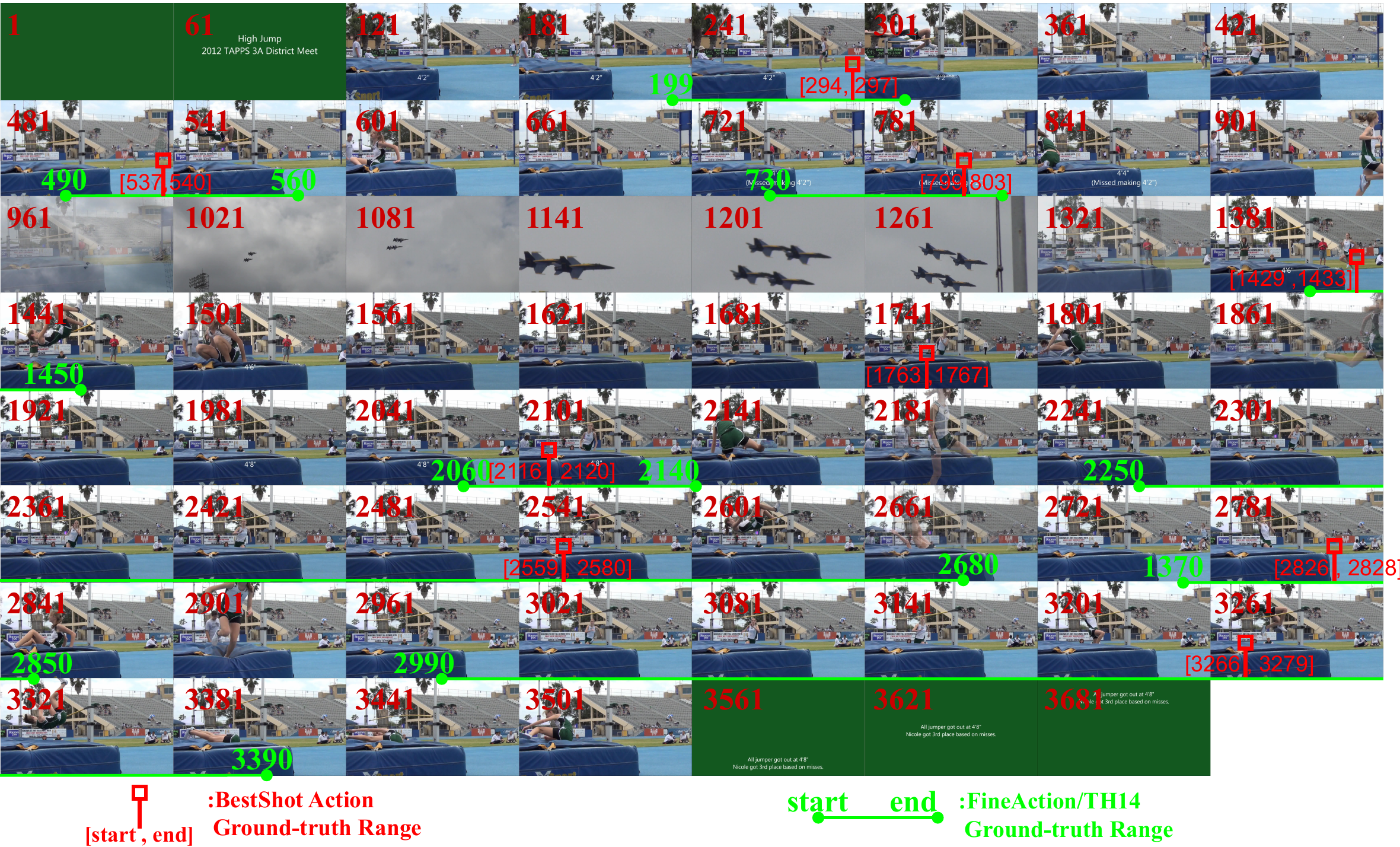} 
\caption{\textbf{Example of BestShot annotation.} This is an example of FineAction with the query \textbf{``high hump.''} BestShot re-annotated the detailed action stage query as \textbf{``In the high jump, during the take-off motion, the body leans backward, preparing to clear the bar.''} and redefined the ground-truth range. We use {\color{red} red} to represent BestShot annotations, and {\color{green} green} for FineAction annotations. The frame number is displayed in the top left corner of the image.}
\label{fig:BestShotExample2}
\end{figure*}

Fig.~\ref{fig:BestShotExample1} shows the difference among MultiSports, QVHighLight, BestShot Benchmark and the video annotation in ShotGPT4o. Fig.~\ref{fig:BestShotExample2} shows the difference between FineAction annotation and BestShot annotation.


\begin{figure*}[!t]
\centering
\includegraphics[width=0.8\textwidth]{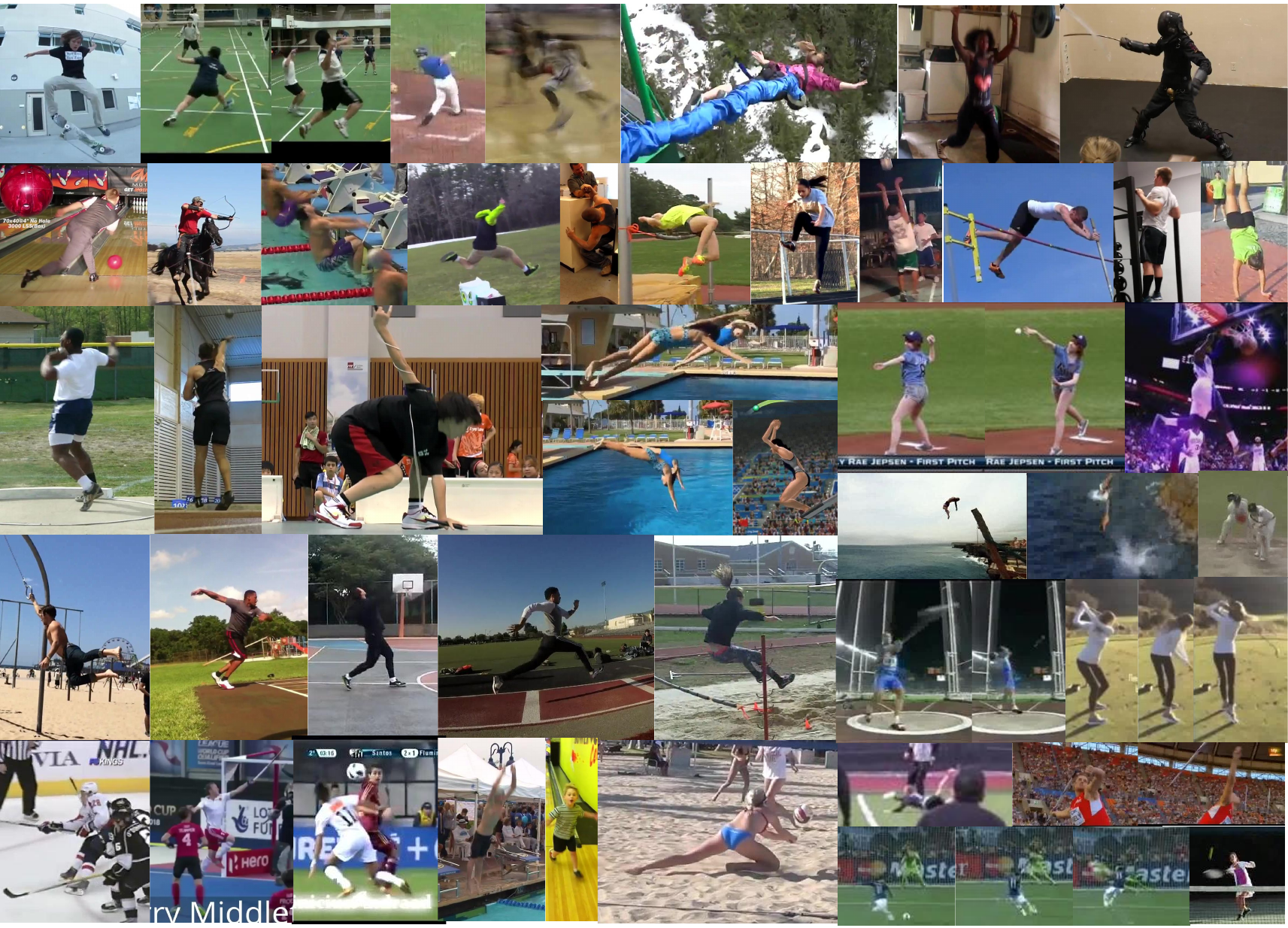} 
\caption{\textbf{Examples of potential highlight frames.}}
\label{fig:expressiveness}
\end{figure*}

\begin{table}[!ht]\small
    \centering
    \begin{tabular}{ccc}
    \hline
        ~ & steps & Charades(IoU\textgreater0.7) \\ \hline
        LITA & 2000 & 4.005 \\ 
        LITA(our data) & 2000 & 6.226 \\ 
        LITA-InternVL & 1500 & 4.178 \\ 
        LITA-InternVL(our data) & 1500 & 7.311 \\ \hline
    \end{tabular}
    \caption{\textbf{Evaluation of original LITA and LITA-InternVL trained with or without our data on Charades.}}
\end{table}

\textbf{Necessity of Potential Highlight Frames} 
Unlike the aesthetic scores in Image Aesthetic Assessment) and the saliency scores in QVHighLight, the expressiveness score is specifically designed to find the key action stage within the entire action, pose with maximum extension, and most exaggerated facial expression. This score is tailored for assessing the saliency of actions in human-centric videos. Since the cost of annotating expressiveness scores is much lower than that of text annotations, its primary contribution in BestShot lies in filtering out most non-highlight frames at a very low cost. The selected frames are shown in Fig.~\ref{fig:expressiveness}. 

\textbf{Necessity of Start-end Labelling} 
Unlike in Moment Retrieval, where a query typically corresponds to only one segment of the video, BestShot videos often contain multiple repeated actions as shown. For instance, the peak of a high jump action may occur multiple times in the video, as illustrated in Fig.~\ref{fig:BestShotExample2}. Therefore, after the text annotation is completed, annotators need to re-annotate all matching start-end frames in the video based on the text.

\textbf{Upper Bound and Human Performance}
We randomly select 5 \% full queries(100 queries) from BestShot Benchmark, and conduct a human performance evaluation to further evaluate the accuracy of benchmark and the upper bound of benchmark scores. Given each clip and the full query, annotators are asked to select three images that best matched the query, and the results will be collected to calculate a top@1 accuracy and top@3 accuracy, resulting in Tab~\ref{tab:human_performance}. 
\begin{table}[]
\centering
\begin{tabular}{cc}
\hline
                 & {Accuracy(Top@1/Top@3)}        \\ \hline
{CLIP}             & {17.9/42.1}          \\
{InternVideo}      & {13.7/34.7}          \\
{InternVL single frame}       & {18.9/46.8}          \\
{HumanPerformance} & {\textbf{87.6/95.2}} \\ \hline
\end{tabular}
\caption{\textbf{Comparisons among existing baseline method and human performance.}}
\label{tab:human_performance}
\end{table}


\section{ShotGPT4o Dataset}

\begin{figure*}[!t]
\centering
\includegraphics[width=0.9\textwidth]{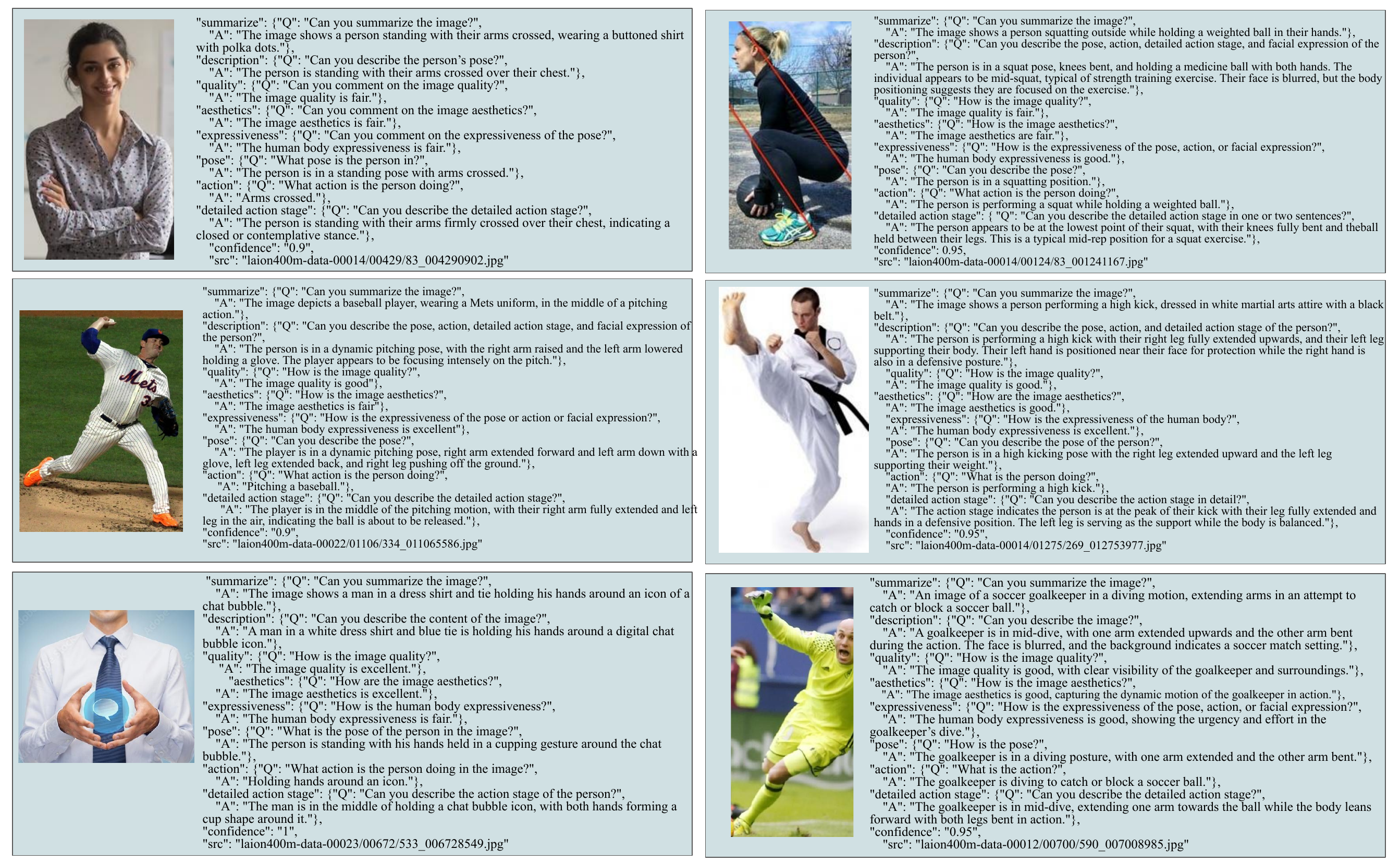} 
\caption{\textbf{More examples of ShotGPT4o Image Q\&As Dataset.} }
\label{fig:ShotGPTimg}
\end{figure*}

\begin{figure*}[!t]
\centering
\includegraphics[width=0.95\textwidth]{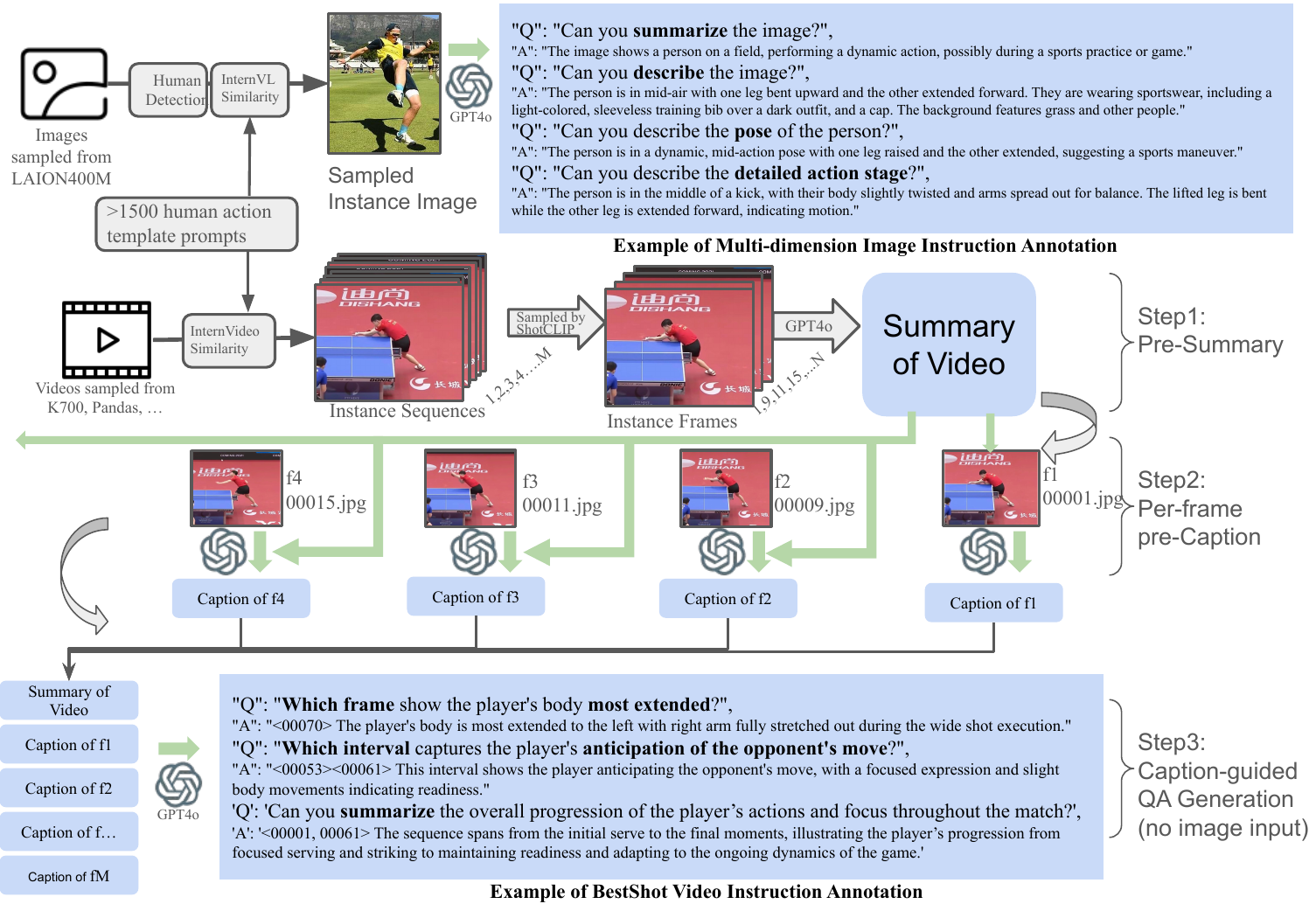} 
\caption{\textbf{Pre-Summary Sliding Shot (PreS3) Captioning Strategy}. We samples videos from K700 and pandas, and images from LAION400M first. For images, we simply ask GPT-4o to generate Q\&As. For videos, we use ShotCLIP to sample frames in videos first. Then we input 10 frames to GPT-4o to generate summary of video. For each sampled frame, both the summary and the individual frame are input into GPT-4 to generate detailed captions for each frame. Finally, the summary and all generated frame captions are used to generate diverse Q\&As.}
\label{fig:ShotGPT2}
\end{figure*}

\begin{figure*}[t]
\centering
\includegraphics[width=0.9\textwidth]{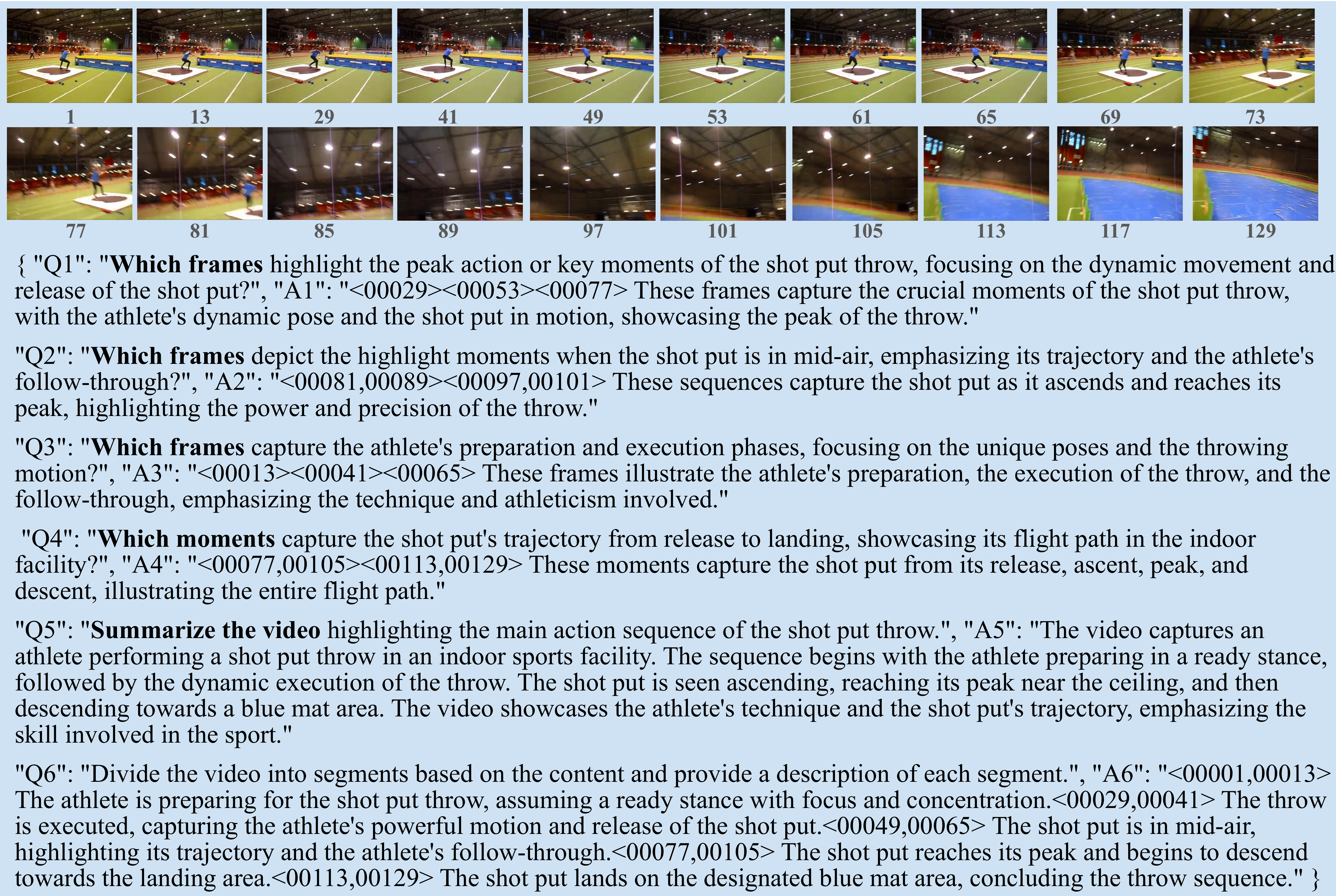} 
\caption{\textbf{More examples of ShotGPT4o Video Q\&As Dataset.} }
\label{fig:ShotGPTvid}
\end{figure*}

\textbf{Image Annotation.}  Fig~\ref{fig:Prompt1} lists the prompt to generate the image Q\&As of ShotGPT4o.

\textbf{Video Annotation.} We provide a detail description of the video annotation method in Fig.~\ref{fig:ShotGPT2}, which we call Pre-Summary Sliding Shot(PreS3) captioning strategy
, aiming to generate Q\&As related to frame retrieval and moment retrieval directly from videos.  This is different from ~\cite{guo2024vtg,huang2024lita} because wo do not convert the existing annotation to Q\&A format, but directly generated new Q\&A with accurate annotation for frame retrieval task and moment retrieval task.
Inspired by ShareGPT4Video’s Differential Sliding-Window Captioning strategy (DiffSW)~\cite{chen2024sharegpt4video}, we adopted a similar approach, first generating frame-by-frame descriptions and then creating video QA based on these descriptions, significantly improving GPT-4's accuracy. However, unlike ShareGPT4Video, we require finer granularity (frame-level) rather than second-level frame sampling to ensure no important frames are missed. 

Therefore, We designed a new captioning strategy, \textbf{P}re-\textbf{S}ummary \textbf{S}liding-\textbf{S}hot (\textbf{PreS3}), with three steps: (1) pre-video summarization, (2) asynchronous frame-by-frame description, and (3) video QA generation. PreS3 offers advantages over DiffSW: (1) Denser and more accurate sampling: ShotVL-Base for dynamic scenes and InternVL for static scenes ensure no key frames are missed. (2) Frame-by-frame descriptions based on summary and single-image inputs: This reduces temporal confusion and hallucinations, making descriptions more robust and asynchronous. (3) QA diversity: Various Q\&As are generated for multiple tasks using open prompts to avoid hallucination.

With such annotation strategy, we sampled 10,000 videos from K700 and generate average 12 Q\&As per video. The Q\&As includes multiple task including frame retrieval, moment retrieval, dense captioning, video summary, temporal reasoning tasks, as shown in Fig.~\ref{fig:ShotGPTvid}.

\textbf{More Examples.}
Fig~\ref{fig:ShotGPTimg} shows more examples of the raw outputs of image Q\&As in ShotGPT4o. The answer of pose, action, detail action stage, description, summary can be treated as caption of image when training ShotVL.  Fig~\ref{fig:ShotGPTvid} show more examples of video Q\&As.

\begin{table}[h]\small
\centering
\begin{tabular}{lll}
\hline 
Pose Description by                & Acc & count     \\ \hline
(a)GPT-4o                          & 63.9\%       & \textbf{18.6M}     \\
(b)ImageSMPLText                   & 87.6\%       & 375K      \\
(c)Human-written                   & \textbf{96.1\%}       & 2K        \\ \hline
Pose Description in ImageSMPLText  & Acc & Frequency \\
(b1)Extrem Pose Subset             & 79.3\%       & 4.0\%    \\
(b2)Truncation or occlusion Subset & 78.4\%       & 8.3\%     \\
(b3)Crop Error(No person)          & 0\%        & 1.3\%     \\ \hline
\end{tabular}
\caption{\textbf{Quantitative comparison of pose description.} Accuracy are compared between pose descriptions generated by Image-SMPLText, GPT-4o and those written manually regarding Image-SMPLText. Accuracy is calculated as the number of incorrect sentences divided by the total number of sentences.}
\label{tab:pose_accuracy}
\end{table}

\section{Image-SMPLText Dataset}

\begin{figure}[t]
\centering
\includegraphics[width=0.9\linewidth]{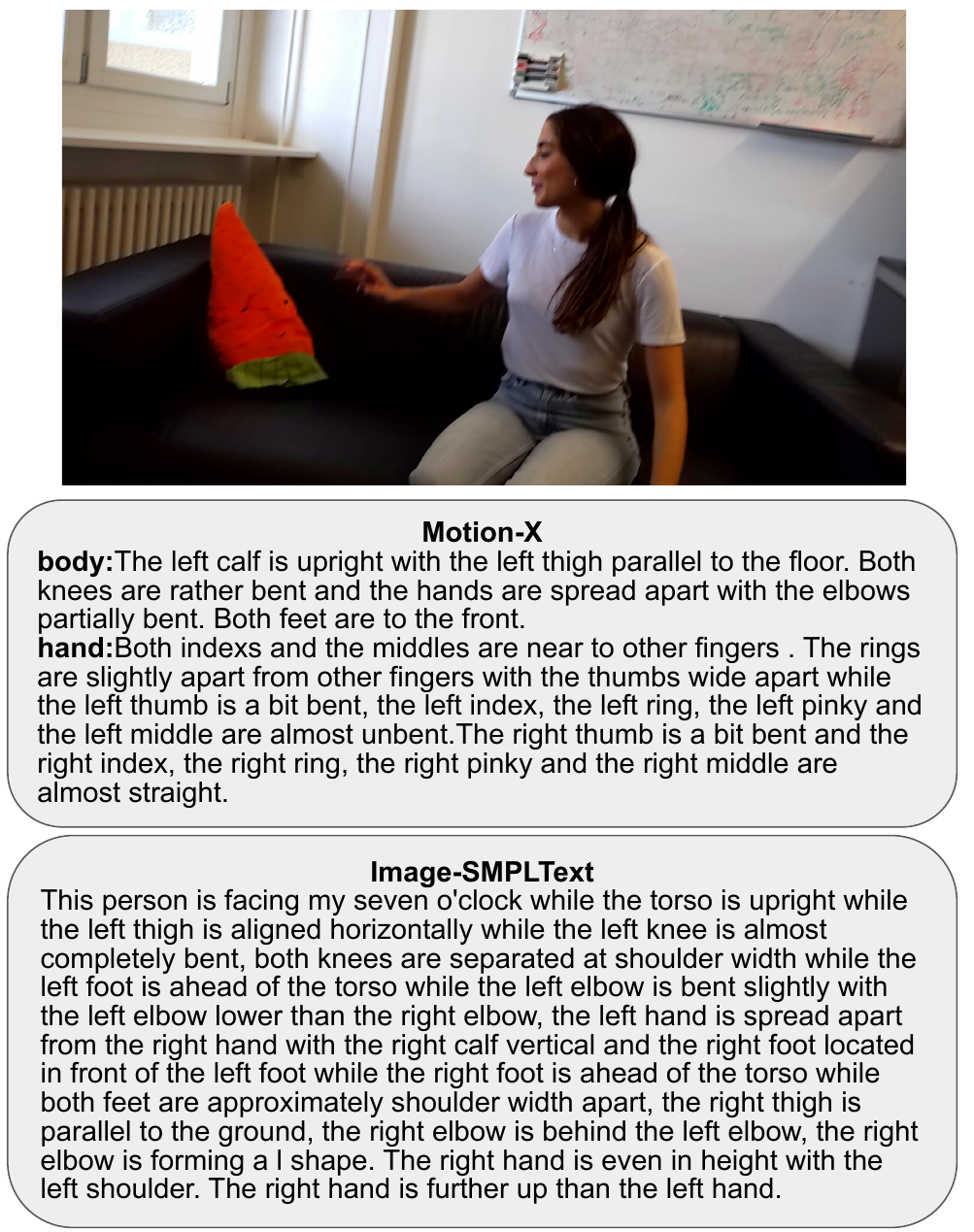} 
\caption{\textbf{Comparison of Motion-X and Image-SMPLText in EgoBody Dataset}. }
\label{fig:posescript_motion_x}
\end{figure}

\textbf{Compare to Motion-X.} In the main paper, we have already compared various configurations of Motion-X and Image-SMPLText, but did not provide a direct visual comparison. Although the datasets sampled by Motion-X and Image-SMPLText differ significantly, both include samples from EgoBody. Therefore, we use EgoBody as an example in Fig.~\ref{fig:posescript_motion_x} to showcase a comparison of the captions they generate.




\textbf{Accuracy Analysis.} 900 pose descriptions (300 for GPT-4o, 300 for Image-SMPLText, 300 for human-written) are randomly sampled to evaluate the accuracy. As shown in Fig.~\ref{tab:pose_accuracy}, the accuracy of pose descriptions generated by GPT-4 is significantly lower than those written by humans. Although the accuracy of Image-SMPLText is still below that of human-written descriptions, it benefits from a larger training sample size. 
We also found that the errors in Image-SMPLText are concentrated in the following categories: (1) Truncation or severe occlusion, where even if the lower body is truncated and not visible, SMPL, being full-body, still generates numerous descriptions about the lower body posture; (2) Extreme poses, such as handstands, lying down, face-up, or side-lying, which may result from SMPL orienting inaccurately under extreme conditions. The lower part of Tab.~\ref{tab:pose_accuracy} lists the accuracy rates for these two conditions.

\begin{figure}[t]
\centering
\includegraphics[width=0.9\linewidth]{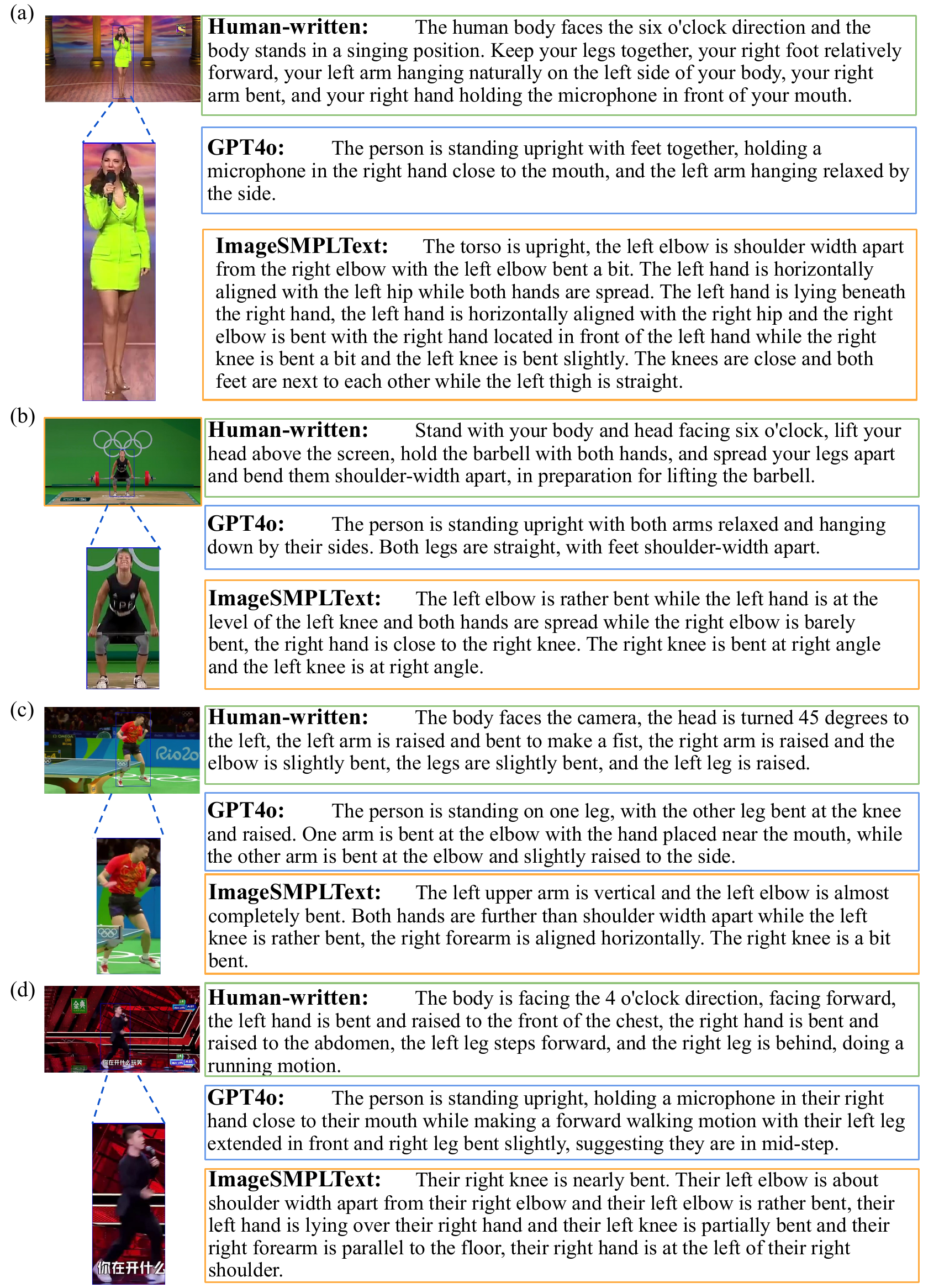} 
\caption{\textbf{More examples of Image-SMPLText Dataset} }
\label{fig:SMPLText}
\end{figure}

\textbf{More Examples.} Fig.~\ref{fig:SMPLText} shows more examples of the Image-SMPLText Dataset. All samples in the dataset are annotated with SMPLText. Additionally, we randomly selected 4,000 and 80,000 samples and labeled them with human-written text and GPT-4o text.

\section{ShotVL}

\begin{figure}[t]
\centering
\includegraphics[width=1\linewidth]{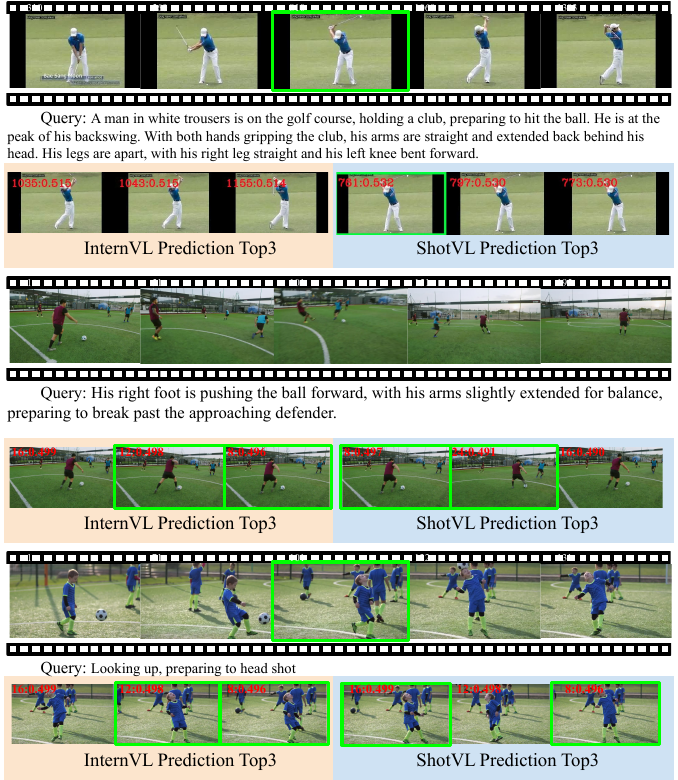} 
\caption{\textbf{Comparison of InternVL and ShotVL in BestShot Benchmark. {\color{red}``Frameid":``similarity score"}} are displayed with {\color{red}red color} on the figure.}
\label{fig:more_compare_shotvl}
\end{figure}

\textbf{Detail of CLIP Benchmark.} 
To evaluate the general performance of ShotVL, we conducted a comparison with InternVL on 24 benchmarks. As illustrated in Table~\ref{tab:clip_benchmark}, ShotVL demonstrated robust classification capabilities on most benchmarks, with only a slight decline in performance. Notably, ShotVL outperformed InternVL on the MNIST, Caltech-101, Flowers-102, and VOC2007 benchmarks.

\begin{table*}[!ht]\small
\setlength{\tabcolsep}{1mm} 
    \centering
    \begin{tabular}{cccccc}
    \hline
        ~ & ImageNet-1K & ImageNet-V2 & ImageNet-A & ImageNet-R & ImageNet-Sketch \\
        ~ & \citeauthor{deng2009imagenet} & \citeauthor{recht2019imagenet} & \citeauthor{hendrycks2021natural} & \citeauthor{hendrycks2021many} & \cite{wang2019learning} \\
        InternVL & \textbf{82.77} & \textbf{76.83} & \textbf{80.29} & \textbf{95.68} & \textbf{73.71} \\ 
        ShotVL & 80.18 & 73.89 & 70.81 & 94.22 & 71.22 \\ \hline
        ~ & ObjectNet & CIFAR-10 & CIFAR-100 & MNIST & Caltech-101 \\ 
        ~ & \citeauthor{barbu2019objectnet} & \citeauthor{krizhevsky2009learning} & \citeauthor{krizhevsky2009learning} & \citeauthor{lecun1998gradient} & \citeauthor{fei2004learning} \\ 
        InternVL & \textbf{73.01} & \textbf{99.43} & \textbf{93.04} & 80.28 & 85.82 \\ 
        ShotVL & 66.75 & 99.00 & 88.70 & \textbf{83.03} & \textbf{85.92} \\ \hline
        ~ & SUN397 & FGVC Aircraft & Country-211 & Stanford Cars & DTD \\ 
        ~ & \citeauthor{xiao2010sun} & \cite{maji2013fine} & \citeauthor{radford2021learning} & \citeauthor{krause20133d} & \citeauthor{cimpoi2014describing} \\ 
        InternVL & \textbf{77.62} & \textbf{51.88} & \textbf{34.09} & \textbf{94.23} & \textbf{70.80} \\ 
        ShotVL & 75.45 & 49.14 & 26.59 & 92.39 & 69.15 \\ \hline
        ~ & Eurosat & FER2013 & Flowers-102 & GTSRB & Pets \\ 
        ~ & \citeauthor{helber2019eurosat} & \citeauthor{goodfellow2013challenges} & \citeauthor{nilsback2008automated} & \citeauthor{stallkamp2012man} & \cite{parkhi2012cats} \\ 
        InternVL & \textbf{75.52} & \textbf{59.25} & 84.19 & \textbf{62.87} & \textbf{96.02} \\ 
        ShotVL & 71.59 & 56.71 & \textbf{84.97} & 59.46 & 93.59 \\ \hline
        ~ & Rendered SST2 & Resisc45 & STL10 & VOC2007 & avg acc1 \\ 
        ~ & \citeauthor{radford2021learning} & \citeauthor{coates2011analysis} & \citeauthor{nilsback2008automated} & \citeauthor{everingham2015pascal} & ~ \\ 
        InternVL & \textbf{67.49} & \textbf{74.41} & \textbf{99.58} & 75.93 & \textbf{77.70} \\ 
        ShotVL & 66.89 & 73.76 & 99.34 & \textbf{80.73} & 75.56 \\ \hline
    \end{tabular}
    \caption{\textbf{A quantitative comparison of the accuracy between InternVL and ShotVL on 24 benchmarks.}}
   \label{tab:clip_benchmark}
\end{table*}

\begin{table*}[!ht]\small
    \centering
    \begin{tabular}{cccccccc}
    \hline
        method & vision encoder & LLM & res. & GQA & POPE & MMVet \\
        ~ & ~ & ~ & ~ & \cite{hudson2019gqa} & \cite{li2023evaluating} & \cite{yu2023mm} \\\hline
        LLaVA-1.5 & CLIP-L-336px & V-13B & 336 & 63.3 & 85.9 & \textbf{35.4} \\ 
        InternVL-Chat-LLaVa & IViT-6B-224px & V-13B & 336 & \textbf{63.9} & \textbf{87.1} & 33.7 \\ 
        ShotVL-Chat-LLaVa & ShotVL & V-13B & 336 & 59.63 & 86.1 & 32.5 \\ \hline
    \end{tabular}
    \caption{\textbf{A quantitative comparison of the accuracy between LLaVa, InternVL-Chat-LLaVa and ShotVL-Chat-LLaVa.} The table outlines the differences among the three models in terms of vision encoder, LLM, and resolution, and presents their performances across various benchmarks.}
   \label{tab:llava_benchmark}
\end{table*}

\textbf{Qualitative Results.} To exhibit ShotVL's impressive zero-shot performance on the BestShot task, Fig.~\ref{fig:more_compare_shotvl} provides examples comparing the retrieval results between ShotVL and InternVL. We uploaded a video and a language query and let ShotVL and InternVL model to retrieve the most relevant frames according to the retrieval settings. 

\section{ShotVL-Chat-LLaVa}
\textbf{Method and Implementation.} To assess the general chat ability of ShotVL, we followed LLaVa's~\cite{liu2023llava} finetuning pipeline replacing the original CLIP encoder with ShotVL. As described in LLaVa, we first aligned ShotVL with LLM, then finetuned both the projector and LLM. During the alignment phase, we incorporated 100K Q\&As based on ImageSMPLText and 200K single-turn image Q\&As from ShotGPT4o. In the subsequent fine-tuning phase, we added an additional 100K Q\&As based on ImageSMPLText and 200K multiple-turn image Q\&As from ShotGPT4o.

\begin{figure}[!t]
\centering
\includegraphics[width=0.95\linewidth]{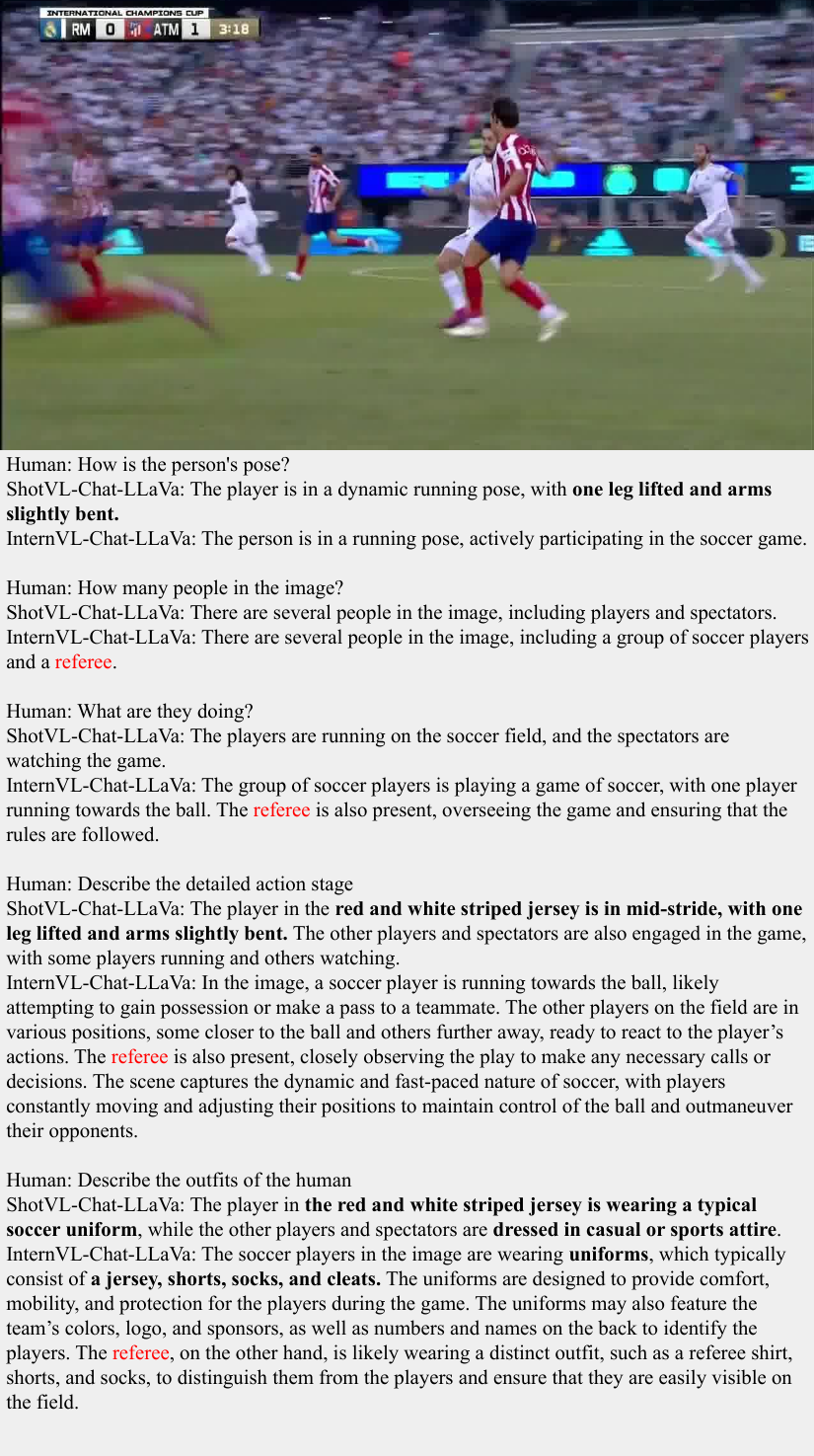} 
\caption{\textbf{Comparion of InternVL-Chat-LLaVa and ShotVL-Chat-LLaVa.} \textbf{Bold} donates ShotVL's detailed pose descriptions and {\color{red}red} donates InternVL's mistakes.}
\label{fig:shotvl-chat}
\end{figure}

\textbf{Qualitative Results.} Tab.~\ref{tab:llava_benchmark} shows the comparison between LLaVa, InternVL-Chat-LLaVa and ShotVL-Chat-LLaVa. Similar to the original ShotVL model, ShotVL-Chat-LLaVa maintained a comparable level of general capability. Fig.~\ref{fig:shotvl-chat} presents an example where we pose questions about the input image. ShotVL-Chat-LLaVa demonstrates greater sensitivity to detailed pose descriptions as we expected.

\begin{figure*}[!t]
\centering
\includegraphics[width=0.8\textwidth]{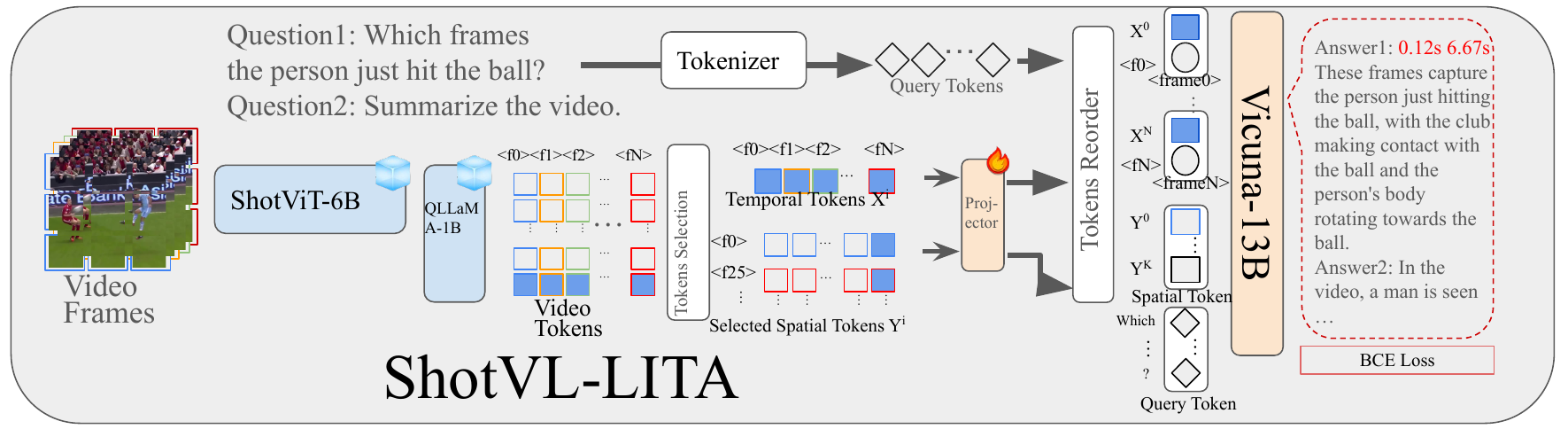} 
\caption{\textbf{Structure of ShotVL-LITA, applying our model based on LITA architecture}}
\label{fig:lita}
\end{figure*}

\begin{figure*}[!t]
\centering
\includegraphics[width=0.9\textwidth]{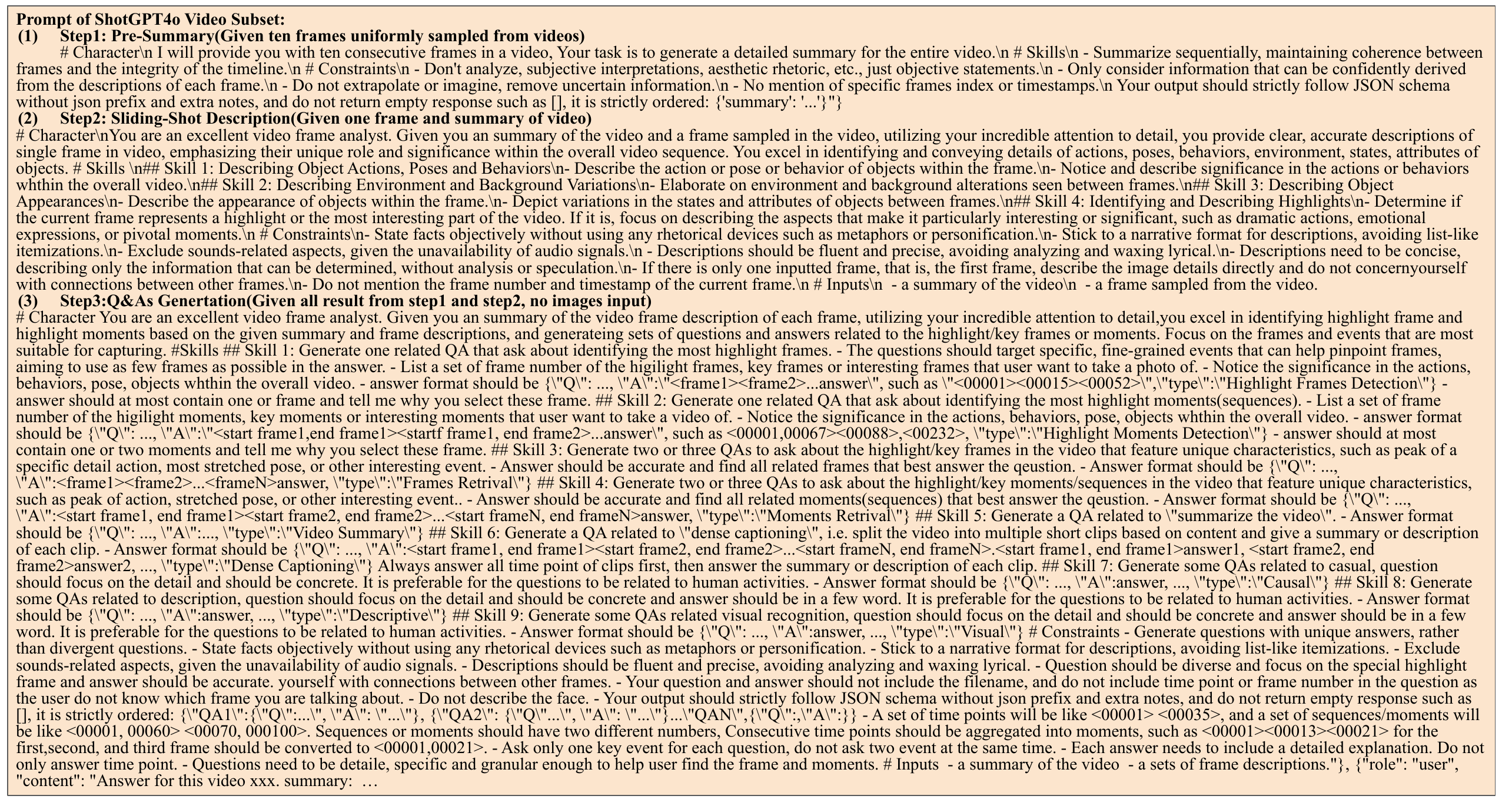} 
\caption{\textbf{Prompt to generate per-frame descriptions and video Q\&A of ShotGPT4o Dataset.} }
\label{fig:Prompt1}
\end{figure*}

\begin{figure}[!t]
\centering
\includegraphics[width=0.95\linewidth]{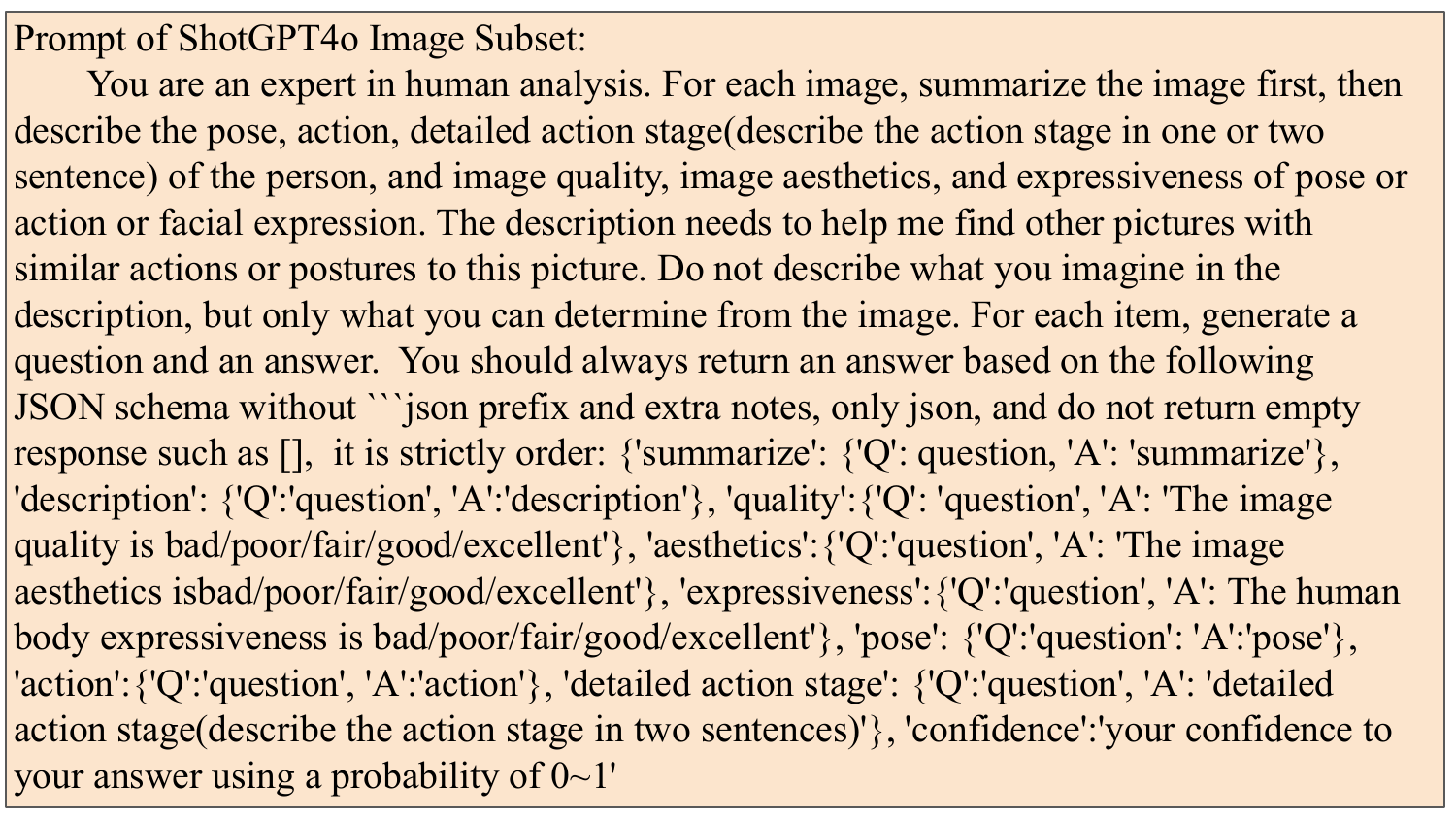} 
\caption{\textbf{Prompt to generate image Q\&A of ShotGPT4o Dataset.} }
\label{fig:Prompt1}
\end{figure}

\section{ShotVL-LITA }
\textbf{Method and Implementation.}
Existing VideoLLM methods with Moment Retrieval capabilities, such as TimeChat~\cite{ren2024timechat}, VTG-LLM~\cite{guo2024vtg}, MR-BLIP~\cite{boris2024surprisingeffectivenessmultimodallarge}, and LITA~\cite{huang2024lita}, have demonstrated the superiority of using LLM as a decoder instead of a Transformer Decoder~\cite{lei2021detecting}. However, these methods are specifically optimized for the Moment Retrieval task and data with a focus on specific scenarios and queries, which may result in huge performance drop on zero-shot BestShot Moment/Frame Retrieval Benchmarks.
We propose crucial modifications to achieve accurate frame localization: Non-equally Sampling, and Interleaving Frame Token. Considering that the modifications are applicable to all aforementioned models, we use LITA as the baseline architecture to explain how to integrate them into such VideoLLM models and their benefits, and call the model in Fig.~\ref{fig:lita} ShotVL-LITA.

\textbf{Effective of ShotGPT4o Dataset.}
To demonstrate the effectiveness of the proposed ShotGPT4o dataset, we extended LITA, an SOTA Video LLM capable of temporal localization, by adding ShotGPT4o to its original training datasets, as detailed per task in Tab.~\ref{tab:task_numbers}. We then trained LITA using the same setup as specified in the original paper, with a learning rate of 2e-5, a batch size of 128, and 4,000 iterations on 16 A800 GPUs.

Upon completing the training, we evaluated both the original and augmented versions of LITA on the BestShot Benchmark, focusing on their frame and moment retrieval capabilities. For reference, we also evaluted VTG-LLM, another SOTA Video LLM capable of moment retrieval, on the same benchmark. However, it is to be noted that VTG-LLM's low score may be attributed to VTG-LLM's outputs being random for some queries, often returning timestamp 000.0 seconds instead of actual time intervals. 

The evaluation results are presented in Tab.~\ref{tab:eval_results}. Since LITA was not initially designed for frame retrieval tasks, the queries in the BestShot Benchmark were modified to match moment retrieval-type questions. The predicted time intervals were calculated using IoU thresholds from 0.3 to 0.7. Although LITA trained with the ShotGPT4o dataset was able to retrieve frames, for an equitable comparison, the predicted frames were expanded to intervals, with pose predictions having a $\pm$4-frame margin, and full and action queries were given a $\pm$6-frame margin. As the results indicate, LITA trained on the augmented data outperformed the original LITA, achieving approximately twice the average score. This highlights the impact of our proposed ShotGPT4o dataset in improving the model's ability to retrieve frames accurately. 
To further illustrate the benefit of our dataset, especially how it helps with the frame localization task, we integrated InternVL, the base model for our proposed ShotVL model, and ShotVL with Vicuna-13B, following the structure of LITA. Minor modifications were made to integrate ShotVL, as mentioned previously. Instead of training on data for all tasks, we focused InternVL-LITA and ShotVL-LITA on the frame retrieval data from ShotGPT4o, comparing their BestShot results with InternVL-LITA trained on all original LITA tasks and datasets. The results, shown in Tab.\ref{tab:eval_shots}, indicate that models trained with ShotGPT4o frame retrieval data outperform the one trained without it. Likely due to a stronger vision model, ShotVL-LITA achieved the highest average scores across all three query types. These results further demonstrated the dataset's ability to improve frame localization capability of models.

\begin{table}[!t]\small
\setlength{\tabcolsep}{1mm} 
\centering
\begin{tabular}{c|cc}
\hline
\textbf{Task} & \textbf{Dataset} & \textbf{Number of Samples} \\ \hline
\multirow{3}{*}{\centering DVC} & ActivityNet Captions & 10009 \\ 
 & YouCook2 & 1237 \\ 
& \cellcolor{yellow}ShotGPT4o & \cellcolor{yellow}4194 \\ \hline
\multirow{2}{*}{\centering Event Localization} & ActivityNet Captions & 10009 \\ 
 & YouCook2 & 1237 \\ \hline
Image Q\&A & LLaVA-150K & 157712 \\ \hline
\multirow{2}{*}{\centering Video Q\&A} & NExT-QA & 37523 \\
 & \cellcolor{yellow}ShotGPT4o & \cellcolor{yellow}16148 \\ \hline
RTL & ActivityRTL & 33557 \\ \hline
\rowcolor{yellow} Moment Retrieval & ShotGPT4o & 12897 \\ \hline
\rowcolor{yellow} Frame Retrieval & ShotGPT4o & 15597 \\ \hline
\end{tabular}
\caption{\textbf{LITA training datasets} with the addition of ShotGPT4o data highlighted in yellow. DVC stands for Dense Video Captioning, and RTL stands for Reasoning Temporal Localization. Datasets are from ActivityNet Captions~\cite{krishna2017densecaptioningeventsvideos}, YouCook2~\cite{Zhou_Xu_Corso_2018}, LLaVA-150K, NExT-QA~\cite{xiao2021next}, and ActivityRTL~\cite{huang2024lita}.}
\label{tab:task_numbers}
\end{table}


\begin{table}[!t]\small
\setlength{\tabcolsep}{0.7mm} 
\centering
\begin{tabular}{l|l|cccccc}
\hline
 & \multicolumn{7}{c}{\textbf{BestShot (IoU)}} \\ \cline{2-8} 
\textbf{Model} & \textbf{Query} & \textbf{0.3} & \textbf{0.4} & \textbf{0.5} & \textbf{0.6} & \textbf{0.7} & \textbf{Avg.} \\
\hline
VTG-LLM & Full & 4.76 & 2.62 & 1.78 & 0.83 & 0.59 & 2.12 \\ 
LITA & Full & 14.05 & 9.20 & 4.98 & 3.48 & \textbf{1.49} & 6.64 \\
LITA$_{ShotGPT4o}$ & Full & \textbf{18.03} & \textbf{14.80} & \textbf{10.95} & \textbf{8.21} & 1.00 & \textbf{10.60} \\
\hline
VTG-LLM & Action & 4.40 & 2.38 & 1.90 & 1.19 & 0.36 & 2.05 \\
LITA & Action & 14.30 & 9.58 & 5.35 & 3.36 & \textbf{1.37} & 6.79 \\ 
LITA$_{ShotGPT4o}$ & Action & \textbf{19.90} & \textbf{16.42} &\textbf{ 13.93 }& \textbf{10.45} & 1.12 &\textbf{ 12.36} \\ 
\hline
VTG-LLM & Pose & 0.12 & 0.12 & 0.00 & 0.00 & 0.00 & 0.05 \\ 
LITA & Pose & 10.70 & 6.42 & 3.21 & 2.14 & 0.83 & 4.66 \\ 
LITA$_{ShotGPT4o}$ & Pose & \textbf{12.96} & \textbf{10.34} & \textbf{7.49} & \textbf{4.52} & \textbf{4.52} & \textbf{7.97 }\\ 
\hline
\end{tabular}
\caption{\textbf{Evaluation of VTG-LLM, original LITA, and LITA trained on ShotGPT4o-extended data on BestShot Benchmark.}}
\label{tab:eval_results}
\end{table}

\begin{table}[!t]\small
\setlength{\tabcolsep}{0.7mm} 
\centering
\begin{tabular}{l|l|cccccc}
\hline
 & \multicolumn{7}{c}{\textbf{BestShot (IoU)}} \\ \cline{2-8} 
\textbf{Model} & \textbf{Query} & \textbf{0.3} & \textbf{0.4} & \textbf{0.5} & \textbf{0.6} & \textbf{0.7} & \textbf{Avg.} \\
\hline
InternVL-LITA & Full & 18.03 & 12.81 & 6.97 & 4.35 & \textbf{2.86} & 9.00 \\
InternVL-LITA$_{FR}$ & Full & 17.79 & 14.30 & 10.32 & 7.34 & 1.12 & 10.17 \\
ShotVL-LITA$_{FR}$ & Full & \textbf{18.16} & \textbf{14.68} & \textbf{10.82} & \textbf{8.08} & 1.24 & \textbf{10.60} \\ \hline
InternVL-LITA & Action & 18.03 & 12.56 & 7.59 & 4.48 & \textbf{2.86} & 9.10 \\
InternVL-LITA$_{FR}$ & Action & 17.79 & 14.05 & \textbf{10.95} & 7.96 & 1.62 & 10.47 \\
ShotVL-LITA$_{FR}$ & Action & \textbf{18.78} & \textbf{15.80} & 10.57 & \textbf{8.83} & 2.11 & \textbf{11.22} \\ \hline
InternVL-LITA & Pose & 12.96 & 9.04 & 4.64 & 3.09 & 1.66 & 6.28 \\
InternVL-LITA$_{FR}$ & Pose & 11.89 & 8.68 & 6.18 & 3.80 & 3.80 & 6.87 \\
ShotVL-LITA$_{FR}$ & Pose & \textbf{17.00} & \textbf{13.32} & \textbf{9.27} & \textbf{4.99} & \textbf{4.99} & \textbf{9.92} \\ 
\hline
\end{tabular}
\caption{\textbf{Evaluation of InternVL-LITA trained on original LITA training data, and InternVL-LITA and ShotVL-LITA trained on Frame Retrieval (FR) data of ShotGPT4o on BestShot Benchmark.}}
\label{tab:eval_shots}
\end{table}

\subsection{Limitation and Future Work}
In this section, we will discuss the limitations of our work and potential directions for future research. We can decompose our work into three parts:

\textbf{Task and Benchmark.} In this work, we introduce the BestShot task and the corresponding benchmark named the BestShot Benchmark. Although we have proven the task and the benchmark's effectiveness, it concentrates on limited categories, mostly sports. To broaden the scope of the task, incorporating additional samples from diverse and previously unexplored scenes is essential.

\textbf{Datasets.} As we discussed in the Image-SMPLText dataset section, we found that the annotations generated by GPT4o show an obvious accuracy decline compared to ImageSMPLText and human-written captions. Meanwhile, 13.6\% of images in our dataset are not suitable for training. In future work, we will optimize the GPT4o-based annotation pipeline and explore how to clean data more effectively. 
For the image Q\&As of ShotGPT4o, we found that even with double the ImageQA training, the performance of ShotVL in BestShot Benchmark did not significantly improve, mainly due to annotation noise and the single-image input training scheme. As for video Q\&As, although we proposed a scalable frame and moment retrieval annotation scheme(PreS3) and generated a small-scale video Q\&As dataset (10,000 videos, 120,000 Q\&A pairs), the frame localization ability of VideoLLM improved after training in ShotGPT4o, but this data scale is still far from sufficient.

\textbf{Model.} Despite the robust performance of ShotVL, its performance did not meet our expectations when transferred to the video modality. The zero-shot performance of ShotVL-LITA, where ShotVL is integrated with an LLM to function as a Video LLM, declined on the BestShot Benchmark, particularly in retrieving frames related to actions. We attribute this decline primarily to an insufficient amount of training data. Moreover, for visual token compression when processing videos in ShotVL-LITA, the same strategy from LITA was employed. However, this compression approach may not be optimal for models specifically designed for frame localization, as details of each frame may be lost during the process. Thus, A better token compression strategy is needed for improving model performance.
\bibliography{main}

\end{document}